\documentclass[10pt,twocolumn,letterpaper]{article}

\usepackage[table,xcdraw,dvipsnames]{xcolor}

\usepackage{iccv}      

\newcommand{\myparagraph}[1]{\vspace{0.4em}\noindent\textbf{#1}}

\newcommand{\gray}[1]{{\color{lightgray}#1}}

\newcommand\first[1]{\cellcolor{red!13}}
\newcommand\second[1]{{\cellcolor{orange!13}}}
\newcommand\third[1]{{\cellcolor{yellow!13}}}

\newcommand\firstB[1]{{\cellcolor{blue!15}}}
\newcommand\firstG[1]{{\cellcolor{green!25}}}
\definecolor{grayline}{HTML}{DDDDDD}
\definecolor{cellhead}{HTML}{F6F6F6}
\newcommand{\mc}{\cellcolor{cellhead}\rule{0pt}{2.5ex}}

\newcommand{\cmark}{\text{\ding{51}}}

\definecolor{iccvblue}{rgb}{0.21,0.49,0.74}
\usepackage[pagebackref,breaklinks,colorlinks,allcolors=iccvblue]{hyperref}
\definecolor{red}{rgb}{1.0	, 0  , 0 }
\usepackage{outlines}
\usepackage{amssymb}
\usepackage{pifont}
\usepackage{multirow}
\usepackage{mathrsfs}
\usepackage{makecell}
\usepackage{hhline}

\usepackage{hhline}
\def\paperID{8044} 
\def\confName{ICCV}
\def\confYear{2025}

\title{Zero-shot Inexact CAD Model Alignment from a Single Image\\[-0.3em]}

\author{
\hspace{-0.7em}
\vspace{0.2em}
Pattaramanee Arsomngern$^{1}$ \quad 
Sasikarn Khwanmuang$^{1}$ \quad 
Matthias Nie{\ss}ner$^{2}$ \quad 
Supasorn Suwajanakorn$^{1}$ \\ 
$^{1}$VISTEC, Thailand \quad
$^{2}$Technical University of Munich, Germany \vspace{-0.1em} \\
\vspace{0.3em}
{\tt\small\url{https://zerocad9d.github.io/}}
}
\begin{document}
\twocolumn[{%
\renewcommand\twocolumn[1][]{#1}%
\maketitle
\begin{center}
\centering
\vspace{-30pt}
  \includegraphics[width=1\linewidth\vspace*{-0.4em}]{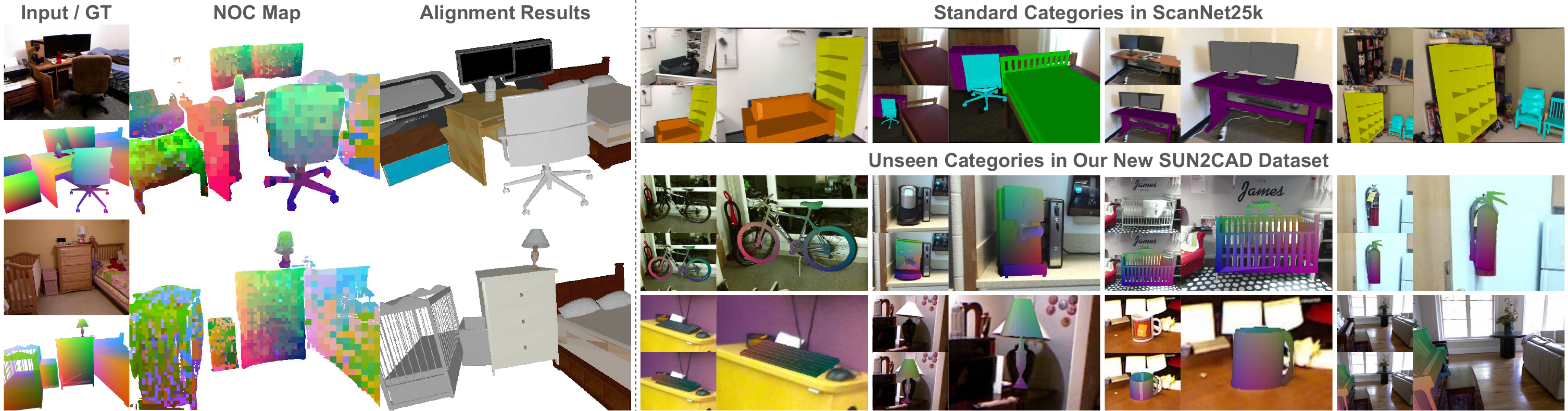}
  \captionof{figure}{We estimate the 9-DoF pose of an inexact CAD model to align it with a target object's pose in a 2D image without training on scene-level pose annotations.
  Our approach can generalize to unseen classes in real images even though it was trained on just 9 classes.}
  \label{fig:teaser}
\end{center}
}]


\begin{abstract}
One practical approach to infer 3D scene structure from a single image is to retrieve a closely matching 3D model from a database and align it with the object in the image. 
Existing methods rely on supervised training with images and pose annotations, which limits them to a narrow set of object categories. To address this, we propose a weakly supervised 9-DoF alignment method for inexact 3D models that requires no scene-level pose annotations and generalizes to unseen categories. Our approach derives a novel feature space based on foundation features that ensure multi-view consistency and overcome symmetry ambiguities inherent in foundation features using a self-supervised triplet loss.
Additionally, we introduce a texture-invariant pose refinement technique that performs dense alignment in normalized object coordinates, estimated through the enhanced feature space.
We conduct extensive evaluations on the real-world ScanNet25k dataset, where our method outperforms SOTA weakly supervised baselines by $+4.3\%$ mean alignment accuracy and is the only weakly supervised approach to surpass the supervised ROCA by $+2.7\%$.
To assess generalization, we introduce SUN2CAD, a real-world test set with 20 novel object categories, where our method achieves SOTA results without prior training on them.
\end{abstract}

\section{Introduction}
\label{sec:intro}
Recovering 3D scene structure from a single image is highly ill-posed, not only due to depth prediction ambiguities but also because large portions of objects are occluded, making full object reconstruction difficult.
One way to address this is by aligning existing 3D models to objects in the input image~\cite{mask2cad,patch2cad,roca,diffcad}.
This approach leverages artist-crafted 3D models, which offer detailed and complete geometry, even for occluded regions, and is particularly well-suited for applications like VR and gaming, where visual realism takes priority over exact geometric fidelity. 

Unlike 6-DoF CAD alignment tasks \cite{foundationpose,megapose,foundpose}, where the 3D model exactly matches the object, this task involves aligning an \emph{inexact} 3D model, retrieved from a database, that may differ in shape or texture or lack texture entirely, to its object's pose in an image. 
In this paper, we propose an approach that eliminates the need for pose annotations to solve this task, enabling generalization to novel objects in unseen categories in a zero-shot manner.

%

Existing methods~\cite{roca,sparc,patch2cad} tackle this problem by relying on extensive 3D supervision from annotated tuples of RGB images, depth maps, CAD models, and 9-DoF poses. 
However, by training on annotated poses from a limited set of categories, they struggle to generalize to unseen categories that differ significantly from the training set~\cite{scannet,scan2cad}. 

To address annotation scarcity, several studies utilize synthetic data for training~\cite{diffcad, foundationpose}.
DiffCAD~\cite{diffcad} relies on the dataset 3D-FRONT~\cite{3dfront}, which provides 9-DoF pose annotations for CAD models in synthetic indoor scenes. 
However, its pose estimator is category-specific and still requires CAD models from that category to synthesize training data, making it difficult to scale and infeasible for unseen categories. 
FoundationPose~\cite{foundationpose} constructs its own synthetic dataset, which covers over 1,000 object categories with diverse textures through augmentation. However, it is designed for 6-DoF tasks with matching textures and models, and its performance drops on inexact matches.
Other 6-DoF studies~\cite{foundpose,freeze} leverage semantic features from 2D foundation models like DINOv2~\cite{dinov2} to establish zero-shot 2D-3D correspondences for pose estimation.
As our study shows, DINOv2-based techniques also struggle in our 9-DoF task with inexact matches or unseen categories.
This performance drop partly stems from key limitations of foundation features inherited by these methods. 
First, foundation features often appear similar in terms of vector distance for symmetrical parts, such as the left and right legs of a chair, as reported by many studies \cite{Zhang2023TellingLF,mariotti2024improving}.
While this property benefits semantic understanding tasks, where both legs can be grouped semantically, our task critically relies on precise differentiation between such parts. Second, these features are sensitive to texture variations~\cite{sddino,freeze}, making matching less consistent, especially for textureless models common in online collections.

To this end, we introduce a technique to enhance foundation features and integrate them into a novel 3D alignment pipeline with a coarse-to-fine estimation scheme: The first step estimates a coarse 9-DoF pose by utilizing a new geometry-aware feature space, derived from DINOv2 features, that is more robust to object part symmetries. The second step refines the pose through dense alignment optimization in a texture-invariant space called the Normalized Object Coordinate (NOC)~\cite{nocs}, for which we also propose a new NOC estimator that generalizes better.
Unlike prior work, which requires real or synthetic training scenes with pose-annotated objects, our pipeline uses only easily accessible front-aligned CAD models for supervision.

In coarse alignment, object pixels and 3D model parts are encoded into a shared feature space, where correspondences are found via nearest neighbors and used to estimate pose with least squares. 
The key challenge is designing an effective feature space and encoder. Our solution trains a small \emph{feature adapter} network that converts foundation features, computed from an image or a 3D model rendering, into custom features. This network enforces multi-view consistency, ensuring features for the same part remain similar across views, while distinguishing features for symmetrical parts that are not well separated in the foundation feature space. 
Leveraging direct access to CAD models~\cite{shapenet}, we formulate these objectives into a self-supervised triplet loss~\cite{facenet}.
This new feature space improves geometric awareness while allowing useful semantics in the foundation features to be retained.

In fine alignment, we use dense image-based alignment to optimize the 3D pose by matching the 3D model's rendering to the input image. However, instead of comparing in RGB space, which is impractical due to mismatched texture, we convert both the input and the rendering into NOC maps~\cite{nocs,6dlearning} for comparison. 
These maps assign pixels from the same object part to a shared normalized 3D coordinate, allowing direct matching.
To predict NOC maps, we leverage our feature space and perform nearest neighbor matching, as in coarse alignment. 
Since nearest neighbor matching is invariant to global scaling and shifting in the feature space, our NOC maps can offer improved robustness to domain gaps and have been found to generalize better to real-world images, even outperforming direct NOC regressors trained on the same synthetic renderings.

We evaluate our method on ScanNet~\cite{scannet} and outperform weakly supervised 9-DoF SOTA, DiffCAD~\cite{diffcad}, by $+4.2\%$ in mean alignment accuracy, the supervised baseline, ROCA~\cite{roca}, by $+2.7\%$, 
and an adapted FoundationPose~\cite{foundationpose} in the inexact 6-DoF setting by $+6.9\%$.
Our features improve DINOv2 by $+4.4\%$, while NOC optimization refines coarse poses by $+4.1\%$.
We also introduce SUN2CAD, an inexact 9-DoF test set with 20 unseen categories, where we surpass the supervised SOTA, SPARC \cite{sparc}, and weakly supervised baselines, achieving state-of-the-art generalization with a large margin $+12.7\%$.
 
To summarize, our contributions are:
\begin{itemize}
    \item A zero-shot single-view 3D alignment approach that handles inexact 3D model matches with state-of-the-art generalization to unseen categories.
    \item A technique to enhance foundation features with object geometry and part symmetry awareness, enabling better 2D-3D part matching.
    \item A technique for predicting NOC maps that enables texture-invariant, smooth dense alignment optimization for fine-grained pose estimation.
   \item A new 9-DoF alignment benchmark, SUN2CAD, featuring 20 new categories not seen in any existing benchmark.
\end{itemize}

\section{Related Work}
\label{sec:related}


\noindent\textbf{Single-view CAD Model Alignment.}
Pose estimator frameworks often integrate multiple techniques in a coarse-to-fine manner, starting with a rough pose estimate and refining it for accuracy~\cite{liu2024deep}. 
Template matching~\cite{megapose,gigapose,ausserlechner2024zs6d} finds an initial pose by comparing the image with multi-view rendering templates. Alternatively, an initial pose can be derived through transformation estimation~\cite{umeyama} using sparse~\cite{matchu} or dense~\cite{nocs,corsetti2024open,shugurov2021dpodv2} 2D-3D correspondences.
Pose regression can be used for both initial prediction~\cite{roca} and refinement~\cite{Li_2024}, but these methods require extensive pose annotations and are limited to known instances or categories.
Recent studies~\cite{foundpose,freeze} address this by exploring training-free pose estimators that leverage features from foundation models~\cite{dinov2}. 
However, these pipelines require exact-match CAD models for texture-dependent refinement, limiting their use with retrieved or user-selected inexact-match CAD models.
In contrast, model-free methods~\cite{onepose,foundationpose} employ neural implicit representations to generate both 3D objects and poses but require multiple reference views and often produce lower-quality 3D models.

To achieve visual realism and support inexact models, supervised end-to-end CAD model retrieval and alignment methods, such as Mask2CAD and Patch2CAD, are introduced ~\cite{mask2cad,patch2cad}. These methods extract features from input images and CAD models by training a feature decoder for shared image-CAD space and pose regression. Instead of using only feature vectors, ROCA~\cite{roca} predicts dense 2D-3D correspondences to improve retrieval and pose estimation, while SPARC~\cite{sparc} refines pose predictions of retrieved CAD models via iterative feature learning. 
However, these studies are still limited by the availability of object categories and pose annotations. 
To address limitations in annotations, DiffCAD~\cite{diffcad} explores training a retrieve-and-align framework on photorealistic scene renderings~\cite{3dfront} using a probabilistic diffusion model. However, the domain gap hinders accuracy on real images, making it necessary to generate multiple outputs to reliably achieve accurate poses. Plus, the availability of photorealistic scene renderings remains limited to a few categories.
%
In contrast, our method supports handling inexact 3D models and requires only synthetic renderings without pose annotations to train, enabling zero-shot pose estimation for unseen categories.

\myparagraph{Vision foundation models.}
Recent advances in computer vision have introduced foundation models trained on large-scale datasets, enabling them to generalize to new tasks without needing task-specific data.
These models serve as either feature encoders or task-specific tools, such as for object detection~\cite{sam,groundingdino}, depth estimation~\cite{depthanything,depth_anything_v2}, or image generation~\cite{stablediffusion,sdxl}.
Feature encoders, such as DINO~\cite{dino,dinov2}, are trained with self-supervised methods~\cite{simclr} on unlabeled data or on text-image pairs~\cite{clip}, providing rich zero-shot features or serving as reliable pre-trained weights.
While DINO and SD~\cite{stablediffusion} have been shown in multiple studies to be effective for zero-shot feature matching, they often lack geometric awareness~\cite{Zhang2023TellingLF} and fail to represent textureless objects. To address this, post-processing algorithms and feature adapters have been proposed to improve accuracy through labels or self-supervision based on 2D spatial location~\cite{mariotti2024improving}. 
In our work, we adapt DINOv2~\cite{dinov2} for zero-shot 2D-3D matching and introduce a feature adapter trained on 3D renderings to enhance geometric awareness.

\begin{figure*}
\centering
\includegraphics[width=1\linewidth\vspace*{-0.4em}]{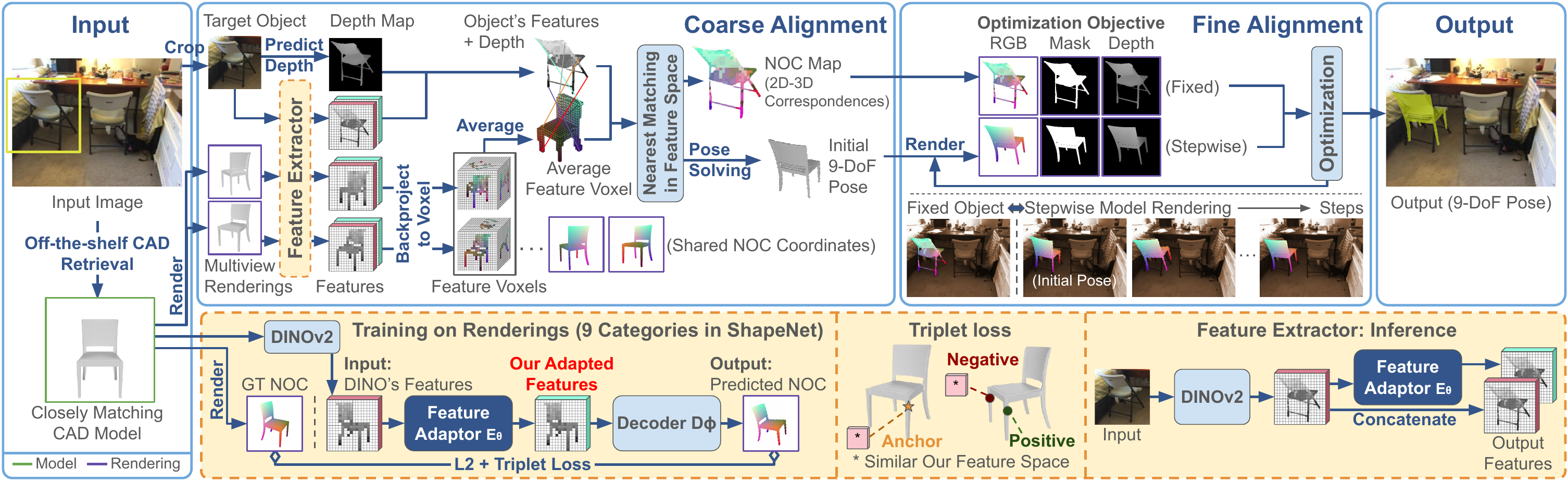}
  \caption{\textbf{Method Overview.} 
  From an image-3D model pair, we first construct 2D and 3D feature grids from DINOv2 and our geometry-aware adapter, trained with 3D self-supervision, then use nearest-neighbor matching to establish correspondences for initial pose solving. Finally, we refine the pose through dense alignment between the predicted NOC map and the 3D model's rendered NOC map.
   }
\vspace{-13pt}

\label{fig:overview}
\end{figure*}


\vspace{-5pt}
\section{Proposed Method}
\label{sec:method}
Given an input image containing a target object, specified by a bounding box, and its closely matching CAD model, our goal is to predict the model's 9-DoF pose in the camera coordinate system, parameterized by a 6D rigid transformation $\mathbf{T}\in \text{SE}(3)$ and an anisotropic scaling vector $\mathbf{s} \in \mathbb{R}^3$. The matching model can be provided by the user or obtained from a retrieval system. In our work, we use ROCA's retrieval system~\cite{roca}. We assume known camera intrinsics. 

This setup challenges traditional CAD alignment techniques~\cite{roca,sparc,patch2cad,mask2cad} as well as recent methods based on foundation features~\cite{foundpose,freeze}, since the retrieved model may differ in shape and texture from the object in the input image. 
To solve this, we propose a novel coarse-to-fine pose estimation method, outlined in Figure~\ref{fig:overview}. We first predict an initial pose by finding 2D-3D correspondence using a novel geometry-aware feature space (Section~\ref{sec:coarse}). Then, we refine this pose through dense alignment (Section~\ref{sec:fine}).

\subsection{Coarse Alignment}
\label{sec:coarse}
To perform coarse alignment, we encode the target object image into a 2D feature map and the CAD model into a 3D feature voxel grid within a shared feature space, then use nearest-neighbors matching to establish 2D-3D correspondences for pose estimation. 
This section first explains how we derive the shared feature space and train its encoder, then how to encode images and 3D models, and finally, how to establish correspondences and solve for pose.




\subsubsection{Geometry-aware feature space}
\label{sec:featureadapter}



A recent trend in zero-shot correspondence matching repurposes foundation models like DINOv2~\cite{dinov2} as image encoders that output 2D feature maps.
This approach has been applied to related tasks, such as 6-DoF pose estimation with an exact model shape and texture~\cite{foundpose,freeze}. However, these features are found to be sensitive to texture variations~\cite{sddino}, which is problematic in our setting where the CAD model may lack matching textures or even be textureless. Furthermore, these features do not prioritize 3D part differentiation or accurately capture geometry and spatial locations, making it difficult to distinguish symmetrical parts, such as chair legs or car wheels. Nonetheless, we hypothesize that these features already contain latent information for predicting part locations, which could be leveraged if restructured.

To this end, we propose training a light-weight MLP feature adapter, $E_{\theta}$, to transform DINO's feature map into a geometry-aware feature map. This adapter is trained on synthetic renderings of CAD models from ShapeNet~\cite{shapenet} with augmentations. 
In this training set, each CAD model is first aligned to a canonical pose, scaled to fit within a unit cube, and rendered from multiple views using an orbiting camera. For augmentations, each rendering is transformed with a conditional diffusion model~\cite{unicontrolnet} to generate variations with added textures and backgrounds, which are added to the training set (Appendix~\ref{supp:augmentation}). The training is done in an encoder-decoder scheme, with $E_{\theta}$ as the encoder and another MLP, $D_\phi$, as the decoder, using two objectives.

\myparagraph{NOC prediction loss.} 
Our first objective encourages our feature to be predictive of 3D locations by solving a coordinate prediction task. Specifically, the decoder's output should predict the Normalized Object Coordinates (NOC) map~\cite{nocs} for each CAD rendering, where each pixel in this map encodes the 3D coordinate of the corresponding point on the model. Let $\mathbf{R}_i$ be a CAD rendering and $\mathbf{N}_i$ its corresponding NOC map, downsampled to the spatial dimension $h \times w$ of the decoder's output. We minimize:
\begin{equation}
\label{eq:reconloss}
\mathcal{L}_{\text{NOC}} = \frac{1}{n\cdot h\cdot w}\sum_{i=1}^n \left\|  D_{\phi}(E_{\theta}(\text{DINO}(\mathbf{R}_i)))-\mathbf{N}_i \right\|^2_2,
\end{equation}
where DINO outputs a 2D feature map from its penultimate layer~\cite{sddino}, and the MLP architectures of $E_{\theta}$ and $D_{\phi}$ process each pixel in the feature map independently. 
We also experimented with architectures that incorporate spatial context, like CNNs and Transformers, but their inductive biases often caused overfitting on limited data---e.g., over-relying on memorized overall shapes rather than utilizing individual DINO features at each pixel to predict NOCs.

\myparagraph{Geometry-consistent triplet loss.}
Another objective ensures that features for the same part appear consistent across viewing angles, while features for parts that are distant in 3D space are distinct. We achieve both tasks with a triplet loss. For each anchor point on a training CAD model:
%
%
%

Its \emph{positive set} gathers model points within a small Euclidean distance of $\tau^{+}_\text{dist}$ from the anchor.

Its \emph{negative set} gathers model points that are at least $\tau^{-}_{\text{dist}}$ away from the anchor \emph{and} whose feature vectors have a cosine similarity above $\tau^{-}_\text{feat}$ with the anchor's feature.

We sample various anchors from various models, along with their positive and negative samples, forming (anchor $\mathbf{a}$, positive $\mathbf{p}$, negative $\mathbf{n}$) into a set $\mathcal{T}$, then compute:
\begin{equation} 
\label{eq:tripletloss}
\mathcal{L}_{\text{triplet}} = \frac{1}{|\mathcal{T}|} \sum_{(\mathbf{a}, \mathbf{p}, \mathbf{n}) \in \mathcal{T}} \left[ d(\mathbf{a}, \mathbf{n}) - d(\mathbf{a}, \mathbf{p}) + \alpha \right]_{+},
\end{equation}
where $d(\mathbf{x}, \mathbf{y})$ computes the cosine similarity between the \emph{encoded features} of model points $\mathbf{x}$ and $\mathbf{y}$, which may come from different rendering views of the same model, and $\alpha$ is a margin threshold used in standard triplet training~\cite{facenet}.

\begin{figure}[!htbp]
\centering
\vspace{-10pt}
  \includegraphics[width=219pt]{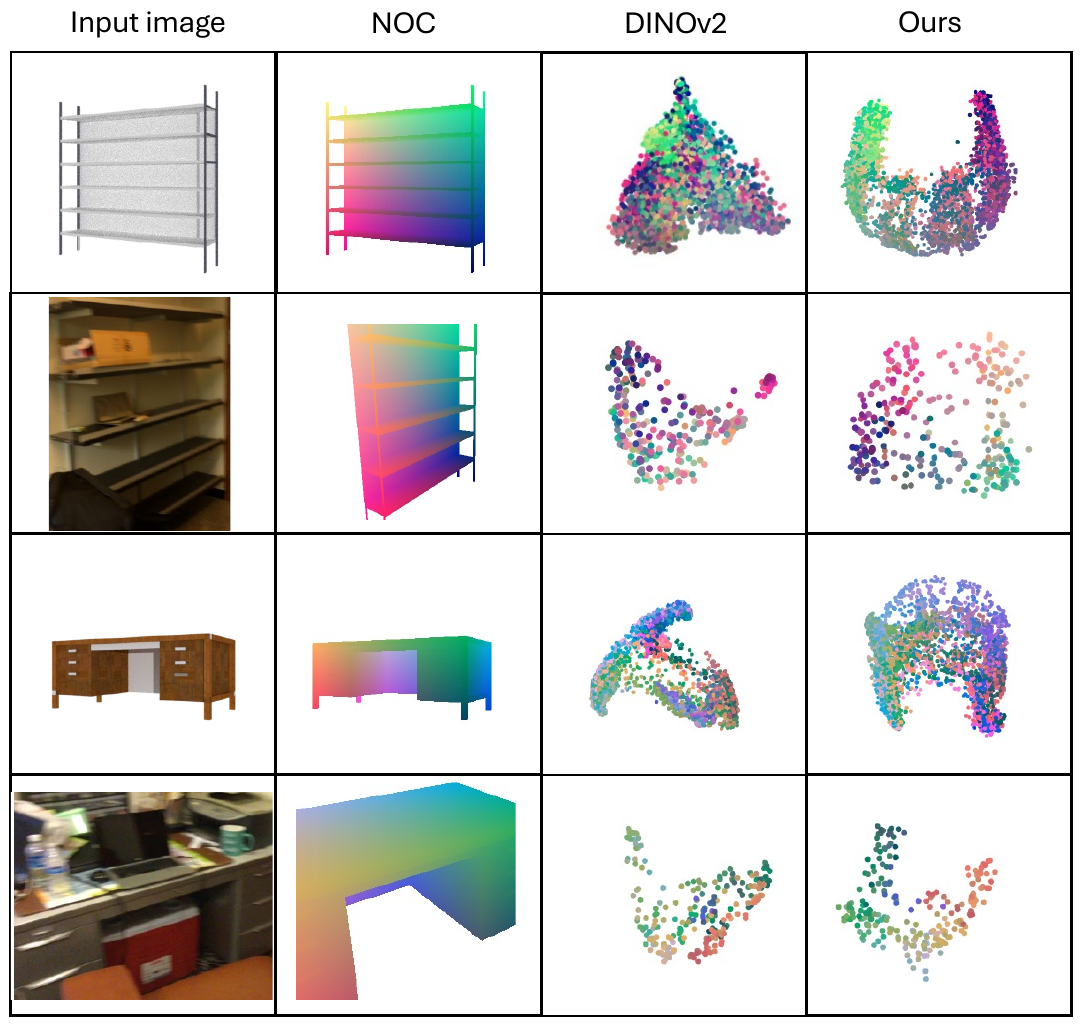}
  \vspace*{-10pt}\caption{\textbf{Visualization of our learned space.} DINOv2 and our geometry-aware features are dimensionally reduced with PCA and color-coded by NOC (see Appendix~\ref{sec:fig_explain}). Nearby object parts are more clearly separated in our features compared to DINOv2.}
  \vspace{-14pt}

  \label{fig:latent}
\end{figure}

\myparagraph{Training.} We train both $E_\theta$ and $D_\phi$, which is later discarded, using the combined loss:
$\mathcal{L}_{\text{adapter}} = (1-\beta)\mathcal{L}_{\text{NOC}} + \beta\mathcal{L}_{\text{triplet}}$, 
where $\beta$ balances the two objectives. These objectives enable $E_\theta$ to become more geometry-aware, as illustrated in Fig.~\ref{fig:latent}, mitigating the 3D perception limitations of prior methods without relying on annotated real datasets.

\subsubsection{Encoding images \& 3D models into shared space}
While the feature from $E_\theta$ enhances geometry-awareness, we found it beneficial to concatenate it with the original DINOv2 features for improved matching. 
In particular, the final feature $E_f(\mathbf{I}) = (1-\omega) \cdot \hat{\text{DINO}}(\mathbf{I}) \oplus \omega \cdot \hat{E}_\theta(\text{DINO}(\mathbf{I}))$, where $\hat{\cdot}$ denotes normalization of the output to unit length.


Given $E_f$, encoding the input image $\mathbf{I}$ into a 2D feature map is simply $E_f(\mathbf{I})$. For the retrieved CAD model, we apply the same function to its multi-view renderings, yielding $E_f(\mathbf{R}_i)$ for each $\mathbf{R}_i$. Using each $\mathbf{R}_i$'s z-buffer and known extrinsics, these features are back-projected into the same 3D world space and stored in a lower-resolution voxel grid (with many voxels empty due to background areas or self-occlusion). Finally, we average across the grids from all $\mathbf{R}_i$ to produce a single feature voxel grid. 
We further apply simple smoothing to the voxel grid, detailed in Appendix \ref{supp:ourimplementation}.

\subsubsection{Feature matching and pose solving.}
\label{sec:posesolving}
For matching, since $E_f(\mathbf{I})$ computes features for the entire input image, including unwanted background elements, we first use SAM~\cite{sam} to segment the object from the given bounding box, then consider only features within the object mask. For each of these features, we find the closest feature in the voxel grid in terms of cosine similarity, resulting in a set of correspondences between 2D feature coordinates and 3D voxel coordinates for initial pose estimation.

Given such correspondences, it is possible to solve for pose using standard techniques such as PnP algorithms~\cite{gao2003complete}. However, this approach can be quite sensitive to scene scale ambiguity. Instead, we leverage a recent metric depth estimator~\cite{depthanything} to predict the depth map of the input image, which lifts the 2D coordinates of each feature at $(u,v)$ in the feature map to 3D $(u,v,d)$. We then convert these to standard $(x, y, z)$ via back-projection and estimate pose from 3D-3D correspondences using a RANSAC-based pose estimator~\cite{ransac,umeyama}. The final coarse pose is estimated using all inlier correspondences, as is standard practice.

In our experiments, we compare our method to an alternative that directly predicts NOC for 2D-3D correspondence. Specifically, each feature in the 2D feature map can use $D_\phi(E_\theta(\mathbf{I}))$, which already encodes normalized 3D coordinates for correspondence. 
We found that our nearest neighbor matching is more robust to distribution shifts, which are particularly significant when training on synthetic data and testing on real images, compared to directly using the NOC output from a neural network, which can be less predictable with out-of-distribution images.

\subsection{Dense Image-based Alignment}
\label{sec:fine}
This step aims to refine the coarse pose using dense image-based alignment, which traditionally optimizes the pose parameters so that the rendering matches the input. To increase invariance to object textures, recent methods like ~\cite{foundpose,freeze} perform the comparison in feature space instead, using DINO's features. While replacing DINO's features with our own in this refinement step readily improves performance (see Section \ref{sec:ablation_pipeline}), we instead propose performing the comparison in NOC space, along with depth and mask spaces, for two reasons. First, each optimization step requires rendering the 3D model in the target space, and rendering it as a NOC map requires only a rasterizer, as opposed to a costly network inference, and consistently yields smooth images. These smooth images help provide stable gradients in dense alignment, where every pixel is considered. 
Second, a NOC map of the input image is inferred only once using nearest neighbor matching in our feature space, providing efficiency, robustness to global shifting and scaling in the feature space, and texture invariance.
To convert the input image $\mathbf{I}$ to a NOC map, denoted by $\mathbf{N}^\mathbf{I}$, we use the same process as before: first computing a feature map $E_f(\mathbf{I})$ and link each feature to its closest 3D voxel, whose location already represents normalized object coordinates (NOC). The $h \times w$ NOC map is upsampled to the $H \times W$ input size via bilinear interpolation. At optimization step $t$, we render the posed 3D model into a NOC map, denoted by $\mathbf{N}^t$, using a differentiable renderer~\cite{pytorch3d}. 
The total loss consists of $\lambda_{\text{NOC-A}} \mathcal{L}_{\text{NOC-A}}(t) + \lambda_{\text{m}}\mathcal{L}_{\text{mask}}(t) + \lambda_{\text{d}}\mathcal{L}_{\text{depth}}(t)$, where $\lambda_{(\cdot)}$ are balancing weights.

\myparagraph{NOC alignment loss} ensures that the predicted NOC matches the rendered NOC from the posed 3D model:
\begin{equation}
\label{eq:fine_noc}
\mathcal{L}_{\text{NOC-A}}(t) =  \frac{1}{ m} \left\| \mathbf{M} \odot (\mathbf{N}^\mathbf{I} - \mathbf{N}^t) \right\|_1,
\end{equation}
where $\odot$ is element-wise multiplication, $\mathbf{M}$ is a binary mask indicating overlapping pixels between $\mathbf{N}^\mathbf{I}$ and $\mathbf{N}^t$, and $m$ is the count of these pixels.

\myparagraph{Silhouette loss} ensures that the silhouettes of the 3D model and the input image match:

\begin{equation}
\label{eq:fine_sil}
\mathcal{L}_{\text{mask}}(t) = \frac{1}{HW} \left\|  \mathbf{S}^\mathbf{I} - \mathbf{S}^t  \right\|_1,
\end{equation}
where $\mathbf{S}^\mathbf{I}$ denotes the object mask computed from SAM (Section \ref{sec:posesolving}), and $\mathbf{S}^t$ is a differentiable soft mask of the 3D model rendered using SoftRasterizer~\cite{pytorch3d,softras}.

\myparagraph{Depth loss} ensures consistency between the depth rendered from the model and the predicted metric depth from the input image.
Unlike NOC and silhouette losses, which focus on 2D projection and are invariant to depth scaling (e.g., enlarging and moving objects away from the camera), this loss is crucial for predicting overall scale and translation, especially along the camera's forward axis.
\begin{equation}
\label{eq:fine_depth}
\mathcal{L}_{\text{depth}}(t) = \frac{1}{m} \left\| \mathbf{M} \odot (\mathbf{D}^\mathbf{I} - \mathbf{D}^t) \right\|_1,
\end{equation}
where $\mathbf{D}^\mathbf{I}$ is the predicted metric depth map from the input image, and $\mathbf{D}^t$ is the rendered depth map.
We minimize the total loss via gradient descent by backpropagating through differentiable renderings to obtain the final pose estimate.

Our use of the three spaces for pose estimation shares some similarities with prior work~\cite{6dlearning}. However, they apply these losses for hypothesis sampling, whereas we use them for dense, differentiable pose optimization.

\begin{table*}[htbp]
\centering
\newcolumntype{x}[1]{>{\centering\arraybackslash\hspace{0pt}}p{#1}}
\resizebox{495pt}{!}{%
\begin{tabular}{l|c|c|x{2cm}|x{2cm}|ccccccccc|cc}
\toprule
 \textbf{Dataset / Metric}&\textbf{Pose} & \textbf{Sup} &\multicolumn{2}{c|}{\textbf{Method}} 
 & \textbf{Bathtub} & \textbf{Bed} & \textbf{Bin} & \textbf{Bkshlf} & \textbf{Cabinet} & \textbf{Chair} & \textbf{Display} & \textbf{Sofa} & \textbf{Table} & \textbf{Avg Cat.$\uparrow$} & \textbf{Avg Inst.$\uparrow$ }  \\ \midrule

\multirow{6}{*}{\makecell[l]{\textbf{ScanNet25k} \\ NMS alignment \\accuracy~\cite{vid2cad}}} &
\multirow{4}{*}{9D} & \cmark & \multicolumn{2}{c|}{ROCA~\cite{roca}}
 &  \second{}22.5    &   10.0  & \first{}\textbf{29.3}    &  \second{}14.2      &   \second{}15.8      &   41.0    &   \first{}\textbf{30.4}   & \third{}15.9  &  14.6    &    \third{}21.5   &   \third{}27.4 
\\
& & \cmark& \multicolumn{2}{c|}{SPARC~\cite{sparc}}&    \first{}\textbf{26.7}     &  \first{}\textbf{25.7}   &   \third{}26.7  &   \first{}\textbf{17.5}     &    \first{}\textbf{23.8}     &    \first{}\textbf{52.6}   &   22.5      &   \second{}32.7   &  \first{}\textbf{17.7}     &   \first{}\textbf{27.3}    &   \first{}\textbf{33.9}\\ 

&& $\star$ &\multicolumn{2}{c|}{FoundationPose~\cite{foundationpose} (for 9D)}
&  \third{}20.0 & \second{}22.9  & \second{}27.6  & 0.9 & 3.1  &  \third{}41.8  &  \third{}23.6  & 15.0  & \second{}17.5  & 19.2  & 25.7
\\

&& $\star$ &\multicolumn{2}{c|}{\textbf{Ours}}
& 16.7 & \third{}18.6  & 22.8  & \third{}12.7  & \third{}9.2  & \second{}49.3  & \second{}24.1 & \first{}\textbf{38.1}  & \third{}16.5  & \second{}23.1  & \second{}30.1\\ \hhline{~---------------} \rule{0pt}{2.3ex}

&\multirow{2}{*}{6D} 
& $\star$ &\multicolumn{2}{c|}{FoundationPose~\cite{foundationpose}}
& \first{}\textbf{22.5} & 21.4  & \first{}\textbf{37.5}  & 6.1  & 5.8  & 44.5  & 30.4  & 29.2  & \first{}\textbf{27.1}  & 24.9  & 31.1
\\
&&$\star$&\multicolumn{2}{c|}{\textbf{Ours} (for 6D)}&
20.8 & \first{}\textbf{25.7}  & 27.6  & \first{}\textbf{19.8}  & \first{}\textbf{22.7}  & \first{}\textbf{56.1}  & \first{}\textbf{51.8}  & \first{}\textbf{45.1}  & 20.1  & \first{}\textbf{32.2}  & \first{}\textbf{38.0}
\\ \hline

\multirow{4}{*}{\makecell[l]{\textbf{ScanNet25k} \\ \textbf{(DiffCAD's split)} \\ Single-view \\ accuracy.~\cite{diffcad}}}
&
\multirow{4}{*}{9D} &$\star$& 
\multicolumn{2}{c|}{\gray{DiffCAD (GT)~\cite{diffcad}}} \rule{0pt}{2.3ex}& -
&\gray{27.1} &\gray{-}& \gray{24.4} & \gray{33.0}&\gray{ 65.9}& \gray{-}& \gray{46.3}& \gray{18.3} & \gray{35.8} &\gray{41.9}
\\
&&$\star$& \multicolumn{2}{c|}{DiffCAD (Mean)~\cite{diffcad}} &-
& 5.2 &- &  1.1   &5.1 &25.2 &-&6.5 & 1.9 &  7.5    &11.7
\\
&&$\star$& \multicolumn{2}{c|}{DiffCAD (Err)~\cite{diffcad}} & -
&\first{}\textbf{7.7} &-&7.6 & 10.4 & 31.1 &-& 15.3 & 4.3 &12.7 &16.7
\\ 
&&$\star$& \multicolumn{2}{c|}{\textbf{Ours} } & -
& 2.8 & -&  \first{}\textbf{8.2}   & \first{}\textbf{12.0} &\first{}\textbf{41.3} &- &\first{}\textbf{21.4} & \first{}\textbf{9.8} &  \first{}\textbf{15.9}    & \first{}\textbf{20.9}\\ \hline
\multicolumn{3}{c|}{\mc \textbf{Ablation Study} } & \mc \textbf{Coarse}     &   \mc \textbf{Fine} & \multicolumn{11}{c}{\mc } 
\\
\hline \rule{0pt}{2.3ex}
\multirow{9}{*}{\makecell[l]{\textbf{ScanNet25k}\\ NMS alignment \\accuracy~\cite{vid2cad}}}
&
\multirow{9}{*}{9D} & 
- &  DINOv2 & - & 14.2 & 5.7  & 6.0 & 8.5  & 6.9  & 34.3  & 9.9  & 24.8  & 7.2  & 13.1  & 18.8\\
&&-&DINOv2&FM &
14.2 &5.7 &9.1 & 9.0 & 7.3 & 33.9 & 9.4 & 23.0 & 7.4 & 13.2 & 18.8
\\
&&-&DINOv2&Ours   & 16.7 & 10.0  & 13.8  & 7.1  & 9.0  & 42.6  & 13.6  & 31.9  & 10.5  & 17.3 &24.2\\
&&$\star$&NOC-S & -
& 5.0 & 1.4  & 1.3  & 4.2  & 3.5  & 34.0  & 17.8  & 12.4  & 0.5  &8.9  & 15.9\\
&&$\star$&NOC-S&Ours &  5.8 & 7.1  & 1.7  & 5.7  & 6.5 & 38.2  & 29.8  & 27.4  &2.5  &13.9  & 19.9  \\ 
&&$\star$& Ours {($\omega$=1)} & - & 15.8 & 4.3 & 6.0 & 10.8 & 6.2 & 38.5 & 17.3 & 23.0 &7.6 &14.4 & 21.0
\\
&&$\star$& Ours {($\omega$=1)} & Ours & 16.7 &11.4 &21.6 &11.8 & 6.2 & 41.5 & 17.3 &23.9 & 10.3 &17.9 & 24.3
\\
&&$\star$& Ours & - & 16.7  & 10.0 & 10.8  & 15.1  & 7.3  &  46.8 & 16.2  & 31.0  & 10.7  & 18.3  & 26.0
\\
&&$\star$&Ours & FM & \textbf{17.5} & 11.4 & 10.3 & 15.1 & 7.7 & 46.1 & 15.7 &30.3 & 11.0 &  18.3 & 26.1
\\
&&$\star$&\textbf{Ours} & \textbf{Ours}& 16.7 & \textbf{18.6}  & \textbf{22.8}  & \textbf{12.7}  & \textbf{9.2}  & \textbf{49.3}  & \textbf{24.1} & \textbf{38.1}  & \textbf{16.5}  & \textbf{23.1}  & \textbf{30.1} \\\bottomrule

\end{tabular}

}
\vspace{-7pt}

\caption{\textbf{Comparison on ScanNet25k~\cite{scan2cad} \& Ablation study.} 
We compare across 3 groups: 9-DoF, 6-DoF, and DiffCAD’s split (Section \ref{sec:mainexp} and \ref{sec:diffcadsplit}), where our method outperforms all non-supervised baselines in mean accuracy.
Supervised, weakly supervised, and unsupervised baselines are marked with `\cmark', `$\star$', and `-', respectively.
We ablate coarse and fine alignment alternatives, detailed in Section \ref{sec:ablation_pipeline}.
} 
\label{tab:newmaintable}
\end{table*}

\begin{table*}[]
\vspace{-10pt}
\addtolength{\tabcolsep}{-0.3em}
\resizebox{\textwidth}{!}{
\begin{tabular}{c|c|c|cccccccccccccccccccc|cc}
\toprule

 

  \textbf{Pose} & \textbf{Sup} &\textbf{Method}  & \rotatebox[origin=c]{90}{\textbf{basket}} & \rotatebox[origin=c]{90}{\textbf{bicycle}} & \rotatebox[origin=c]{90}{\textbf{blender}} & \rotatebox[origin=c]{90}{\textbf{broom}} & \rotatebox[origin=c]{90}{\textbf{clock}} & \rotatebox[origin=c]{90}{\textbf{coffmkr}} & \rotatebox[origin=c]{90}{\textbf{crib}} & \rotatebox[origin=c]{90}{\textbf{fireext}} & \rotatebox[origin=c]{90}{\textbf{keybrd}} & \rotatebox[origin=c]{90}{\textbf{ladder}} & \rotatebox[origin=c]{90}{\textbf{lamp}} & \rotatebox[origin=c]{90}{\textbf{mug}} & \rotatebox[origin=c]{90}{\textbf{piano}} & \rotatebox[origin=c]{90}{\textbf{printer}} & \rotatebox[origin=c]{90}{\textbf{remote}} & \rotatebox[origin=c]{90}{\textbf{shoe}} & \rotatebox[origin=c]{90}{\textbf{phone}} & \rotatebox[origin=c]{90}{\textbf{oven}} & \rotatebox[origin=c]{90}{\textbf{vase}} & \rotatebox[origin=c]{90}{\textbf{bottle}} & \textbf{Cat.$\uparrow$} & \textbf{Inst.$\uparrow$}         \\
 
&& &\#7&\#14&\#7&\#2&\#13&\#19&\#18&\#15&\#66&\#4&\#132&\#59&\#18&\#92&\#8&\#3&\#47&\#14&\#9&\#3&\#20&\#550\\
\midrule
\multirow{3}{*}{9D} & \cmark & SPARC~\cite{sparc} & 
14.3 & 7.1 &0.0 &\first{}\textbf{50.0} & 0.0 & 0.0 &0.0 &0.0 &0.0 &0.0 &3.0 &0.0&27.8 & 14.1 &0.0 &0.0 &0.0 &0.0 &\first{}\textbf{22.2} &0.0 & 6.9 & 4.9
\\  
 &-& DINOv2
 & 0.0& \first{}\textbf{28.6} & \first{}\textbf{14.3} & \first{}\textbf{50.0} & \first{}\textbf{7.7} & 10.5 &0.0 & \first{}\textbf{13.3} &3.0 & 0.0 &5.4 &0.0 &44.4 &7.6 & 0.0 & \first{}\textbf{33.3} &6.4 & 7.1 & 0.0 & 0.0 & 11.6 & 7.3   \\  
 &$\star$ & \textbf{Ours} &  \first{}\textbf{42.9}&21.4&\first{}\textbf{14.3}&\first{}\textbf{50.0}&\first{}\textbf{7.7}&\first{}\textbf{21.1}&\first{}\textbf{16.7}&6.7&\first{}\textbf{24.2}&\first{}\textbf{25.0}&\first{}\textbf{6.1}&\first{}\textbf{10.2}&\first{}\textbf{50.0}&\first{}\textbf{25.0}&\first{}\textbf{25.0}&\first{}\textbf{33.3}&\first{}\textbf{8.5}&\first{}\textbf{57.1} & 11.1 &\first{}\textbf{33.3}  &\first{}\textbf{24.5} &\first{}\textbf{17.6}   \\  \midrule
 \multirow{2}{*}{6D} & $\star$ & FoundationPose~\cite{foundationpose}
 &  28.6 & \first{}\textbf{50.0} & \first{}\textbf{14.3} & \first{}\textbf{50.0} & 7.7 & 5.3 &11.1 & \first{}\textbf{33.3} & 33.3 &25.0 & 23.5 & 15.3 & 11.1 & 7.6 & \first{}\textbf{25.0} & \first{}\textbf{0.0} & \first{}\textbf{29.8} & 21.4 & 0.0 & 33.3 &21.3 & 20.4 \\  
& $\star$ & \textbf{Ours} (for 6D) & \first{}\textbf{42.9} & \first{}\textbf{50.0}& \first{}\textbf{14.3} & \first{}\textbf{50.0} &\first{}\textbf{23.1} &\first{}\textbf{36.8} &\first{}\textbf{16.7} &20.0 & \first{}\textbf{39.4} & \first{}\textbf{50.0} & \first{}\textbf{30.3} & \first{}\textbf{20.3} &\first{}\textbf{61.1} & \first{}\textbf{27.2} &\first{}\textbf{25.0} &\first{}\textbf{0.0} &13.8 &\first{}\textbf{78.6} &\first{}\textbf{44.4} & \first{}\textbf{66.6} & \first{}\textbf{35.5} &\first{}\textbf{30.7}  \\  
 \bottomrule
\end{tabular}
}
\vspace{-7pt}

\caption{\textbf{Comparison on unseen categories in SUN2CAD.} We report single-view accuracy~\cite{diffcad}, and achieve the highest mean accuracies.} 
\vspace{-10pt}
\label{tab:sun2cad}
\end{table*}
 
\section{Experiments}
\label{sec:expresults}
We compare our method against two supervised baselines, SPARC~\cite{sparc} and ROCA~\cite{roca}, and two weakly supervised baselines, DiffCAD~\cite{diffcad} and FoundationPose~\cite{foundationpose}, on inexact-CAD pose estimation. 
SPARC is the current SOTA in 9-DoF pose refinement; however; it requires category-specific median scaling values,
which are unavailable in other competitors' settings.
Thus, we also include ROCA, the second-best supervised 9-DoF baseline, which predicts pose from scratch, similar to our setting. Among weakly supervised baselines, DiffCAD is the SOTA in 9-DoF alignment but is limited to training categories. 
In contrast, FoundationPose represents the best open-source pose estimator that supports unseen categories and textures but estimates only 6-DoF poses without 3-DoF scaling. 

Due to differences in setup and requirements, not every baseline can be evaluated in all scenarios: We compare against SPARC, ROCA, and FoundationPose on ScanNet25K~\cite{scannet} (Section~\ref{sec:mainexp}); DiffCAD on its own test subset of ScanNet25K (Section~\ref{sec:diffcadsplit}); SPARC and FoundationPose on our new SUN2CAD dataset with unseen categories. 
Finally, we conduct ablation studies on our coarse-to-fine pipeline and hyperparameters (Sections~\ref{sec:ablation_pipeline} and \ref{sec:ablation_params}).
See Appendices~\ref{supp:moreexp}$-$\ref{supp:failurecase} for additional results and failure cases.


\myparagraph{Implementation details.}
\label{sec:implement}
Our feature adapter $E_{\theta}$ uses a 2-layer MLP, and the decoder $D_{\phi}$ uses a 1-layer MLP. Both are trained from scratch using the AdamW optimizer~\cite{adamw}.
%
%
For training data, following Scan2CAD~\cite{scan2cad}, we use 9 categories in ShapeNet~\cite{shapenet} dataset to render and augment templates, resulting in 300k images. We use DINOv2- ViT-L~\cite{dinov2} to create input feature maps with the size of $1024$ per patch.
The dense alignment is implemented using PyTorch3D~\cite{pytorch3d} differentiable rendering and optimized with Adam.
We use DepthAnything~\cite{depthanything} as a metric depth estimator, which was fine-tuned on the training data of each benchmark dataset.
All experiments were conducted on a single V100 GPU.
See Appendix~\ref{supp:ourimplementation} for additional details.

\subsection{Comparison on ScanNet25k Dataset}
\label{sec:mainexp}

Following previous studies~\cite{roca,sparc,mask2cad}, we evaluate performance on ScanNet25k~\cite{scannet} using pose annotations from Scan2CAD~\cite{scan2cad} and the standard accuracy metric~\cite{vid2cad,roca,sparc}.
The dataset provides \emph{scene videos}, where each object appears in multiple frames, totaling 20K training and 5K test images. The accuracy metric first applies non-maximum suppression (\emph{NMS})~\cite{vid2cad} to select the pose prediction with the best ROCA's retrieval confidence score across frames, then computes the percentage of predictions with translation, rotation, and scaling errors of $\leq 20$ cm, $\leq 20^\circ$, and $\leq 20\%$ relative to the ground truth. Despite being designed for video-based alignment~\cite{vid2cad}, this NMS-based metric is adopted by single-view methods~\cite{roca,sparc}.
\myparagraph{9-DoF ROCA~\cite{roca} and SPARC~\cite{sparc}:}
As seen in Table~\ref{tab:newmaintable}, our method is the only weakly supervised method to surpass ROCA~\cite{roca} in the average category-wise and instance-wise NMS scores by $\{+2.7\%,+1.6\%\}$. 
SPARC~\cite{sparc} achieves higher average scores but lower scores on Display and Sofa.
Note that SPARC requires per-category median scales for pose initialization, which alone yield $65.2\%$ scaling accuracy without further processing, and cannot handle unseen categories (Table~\ref{tab:sun2cad}).
We observe that our underperforming categories (e.g., table, bathtub, bed) often face large occlusions and cropping, affecting scale and rotation estimates, whereas supervised methods can be less impacted by learning from similar distributions.
(Appendix~\ref{supp:moreexp}).

\myparagraph{6-DoF FoundationPose~\cite{foundationpose}:}
For a fair comparison, we evaluate under two setups, shown in Table~\ref{tab:newmaintable}: (1) An inexact 6-DoF estimation task, which is closest to the FoundationPose's original setup. In this setup, we adapt our method by fixing the scales to the ground truth values used by FoundationPose. (2) An inexact 9-DoF estimation task, where we adapt FoundationPose and provide it with the scales predicted by ROCA ($76.8\%$ accuracy in ScanNet25k~\cite{scannet}). We provide these setups for reference only, as their method is not designed for 9-DoF or inexact match tasks.


The adapted 6-DoF version of our method outperforms FoundationPose in 6 out of 9 classes ($\{+7.3\%,+6.9\%\}$ mean accuracies), and our method also outperform its adapted 9-DoF version in 5 out of 9 classes. 
We observe that their method struggles with rotation in large objects, such as bookshelves and cabinets, which differ from the smaller objects FoundationPose was trained on.

\subsection{Comparison on ScanNet25k---DiffCAD's split}\label{sec:diffcadsplit}
We compare our method to the SOTA 9-DoF weakly supervised method, DiffCAD~\cite{diffcad} in Table~\ref{tab:newmaintable}. 
DiffCAD was trained on six categories from 3D-FRONT~\cite{3dfront} and tested on their own subset of ScanNet25K with the same categories in their paper. DiffCAD's metric does not apply NMS as used with ScanNet25K and involves generating multiple hypotheses and uses the one closest to the ground truth to report scores (DiffCAD(GT) in Table~\ref{tab:newmaintable}; reproduced from official code for reference only). However, our problem setup does not assume access to the ground-truth pose for such hypothesis selection, thus we instead compute two other metrics. (1) DiffCAD (Err), which selects the best hypothesis based on projection errors from its 2D-3D correspondences and the solved pose. (2)
DiffCAD (Mean), which computes the mean error across all hypotheses.
For a fair comparison, our method uses a non-fine-tuned depth estimator since DiffCAD was not trained on ScanNet's depth maps, and we use DiffCAD's own test set for evaluation.  
Our method surpasses both DiffCAD (Err) and DiffCAD (Mean) in 5 of 6 classes, achieving $\{+3.3\%,+4.3\%\}$ and $\{+8.4\%,+9.2\%\}$ mean accuracies.

\begin{figure}[htbp]
\centerline{\includegraphics[width=236pt]{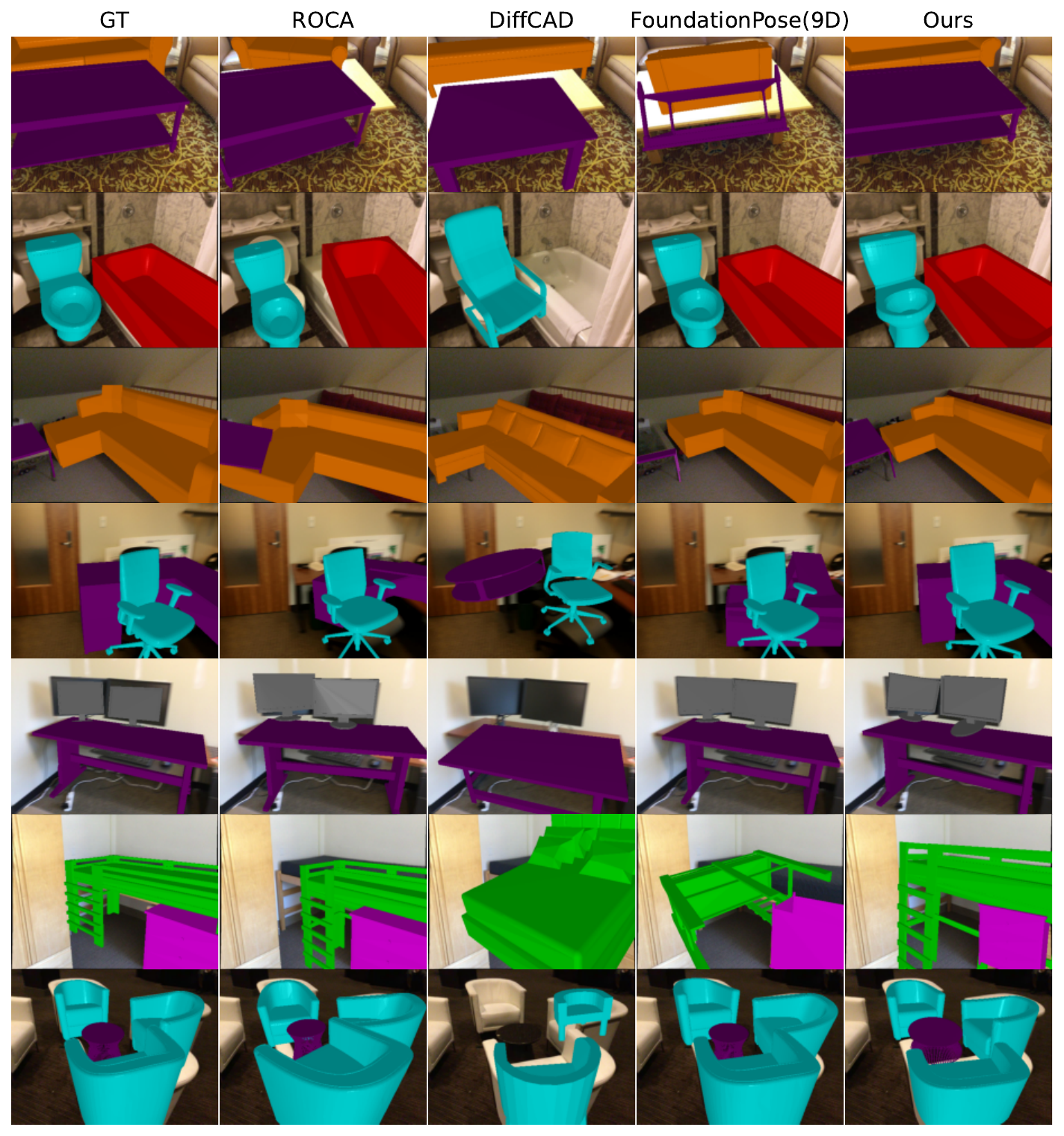}}
\vspace{-10pt}
\caption{\textbf{Qualitative results in ScanNet25k dataset.}}
\label{fig:mainresults}
\vspace{-10pt}
\end{figure}

\subsection{Comparison on SUN2CAD Dataset}
\label{sec:newdatasetexp}
To assess zero-shot capability on unseen categories, we introduce SUN2CAD, a new inexact 9-DoF test set that extends beyond the nine categories in ScanNet25k~\cite{scannet}. SUN2CAD comprises 550 samples across 20 categories, featuring diverse shapes and sizes, including bicycles, pianos, fire extinguishers, and lamps. 
Annotations are derived from 3D bounding boxes in SUN RGB-D~\cite{sunrgbd} and further refined through manual adjustments to improve accuracy.

\begin{figure}[htbp]
\vspace{-8pt}
\centerline{\includegraphics[width=236pt]{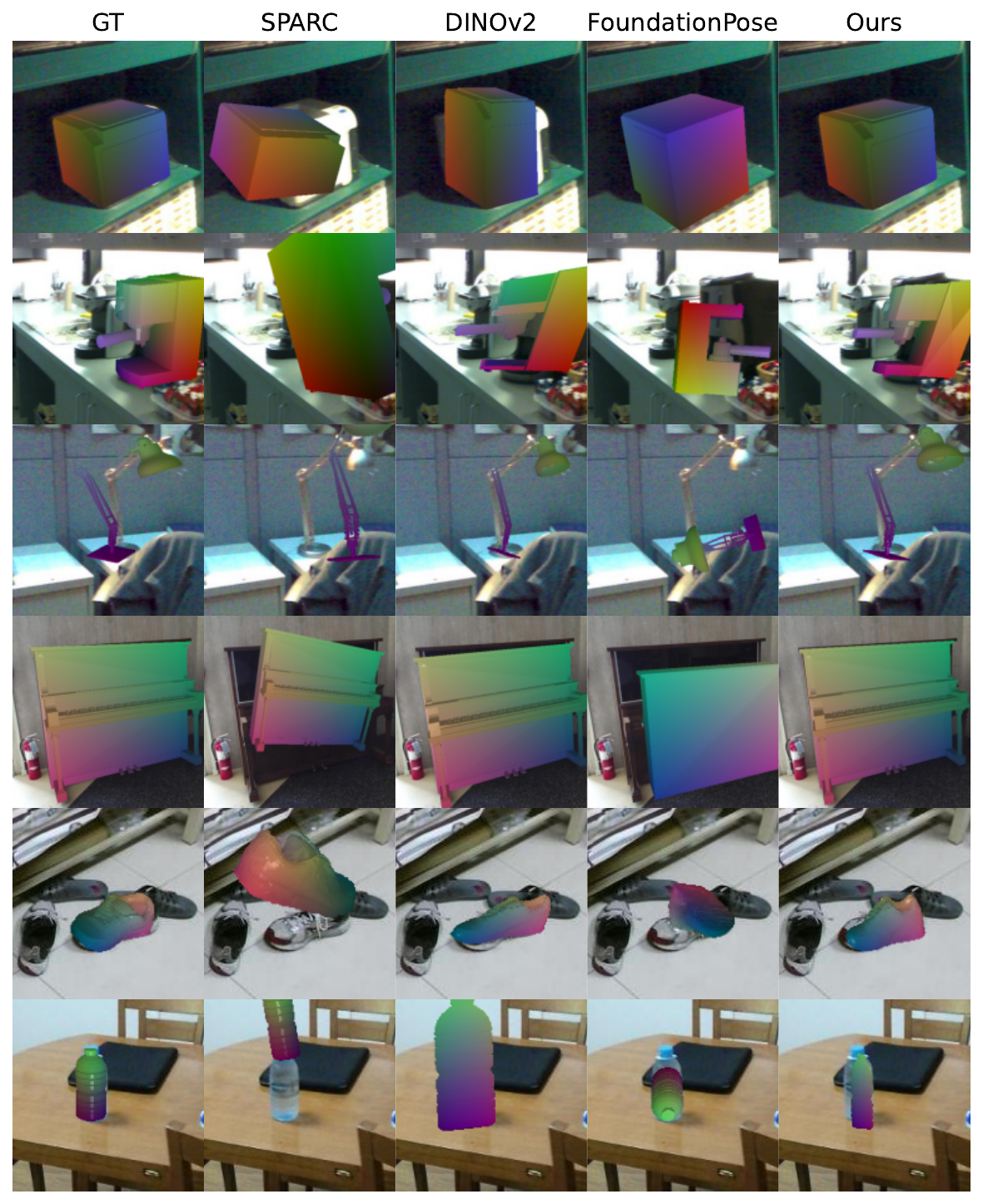}}
\vspace{-10pt}
\caption{\textbf{Qualitative results in SUN2CAD dataset.}}
\label{fig:mainsun}
\vspace{-10pt}
\end{figure}

Table~\ref{tab:sun2cad} and Fig.~\ref{fig:mainsun} compare our method with competitors. Note that we can not evaluate against DiffCAD~\cite{diffcad} because its category-specific pose estimators are incompatible with unseen categories.
While SPARC~\cite{sparc} performs well in seen categories, it struggles to generalize to SUN2CAD (see Appendix~\ref{supp:moreexp} for further analysis). Compared to DINOv2, known for its zero-shot semantic understanding~\cite{dinov2}, our method achieves superior results, suggesting zero-shot capabilities of geometry-aware features. We outperform SPARC in 18/20 and DINOv2 in 14/20 categories, with mean accuracy gains of $\{+17.6\%, +12.7\%\}$ and $\{+12.9\%, +10.3\%\}$, respectively.
In the inexact 6-DoF setting, where ground-truth scales are provided to all competitors, we surpass FoundationPose~\cite{foundationpose}, which is specialized in novel but exact object pose estimation, in 13/20 categories and $\{+14.2\%, +10.3\%\}$ gains.

\subsection{Ablation Study on Coarse-to-fine Pipeline}\label{sec:ablation_pipeline}
We evaluate our pipeline against combinations of coarse and fine alignment alternatives in Table~\ref{tab:newmaintable}. 

\myparagraph{Coarse alignment.}
We test three alternatives:
(1) Using DINOv2~\cite{dinov2} features for nearest neighbor matching instead of our fused geometry-aware features (Section~\ref{sec:posesolving}).
(2) Using only geometry-aware features without DINOv2 ($\omega$=1).
(3) Predicting NOC maps directly from DINOv2 features using a neural network trained on the same synthetic data as our method (denoted by NOC-S in Table~\ref{tab:newmaintable}). See Appendix~\ref{supp:nocs_predictor} for additional details.

\myparagraph{Fine alignment.}
We test two alternatives: (1) No refinement. (2) Performing dense alignment using features used in the coarse step, as proposed by FoundPose~\cite{foundpose} (denoted as FM in the ``Fine'' column). Note that this is a reimplementation, as FoundPose does not provide source code.

The results using only coarse alignment show that our fused features (Ours,$-$) improve accuracy from using pure DINOv2 features (DINOv2$,-$) by $\{+5.2\%,+7.3\%\}$. 
Pure adapted features (Ours ($\omega$=1), $-$) also outperform DINOv2 but underperform our fused version.
When all methods use fine alignment, (Ours, Ours) clearly outperforms (DINOv2, Ours) and surpasses (NOC-S, Ours) in 8 out of 9 classes and in mean accuracy, highlighting the effectiveness of our geometry-aware features.
For fine alignment, our NOC dense optimization improves accuracy compared to both using no refinement $\{+4.8\%,+4.1\%\}$ and using dense alignment based on features ($\cdot,$ FM) by $\{+4.8\%,+4.0\%\}$. 

\subsection{Ablation Study on Hyperparameters}
\label{sec:ablation_params}

\begin{figure}[htbp]
\centerline{\includegraphics[width=220pt]{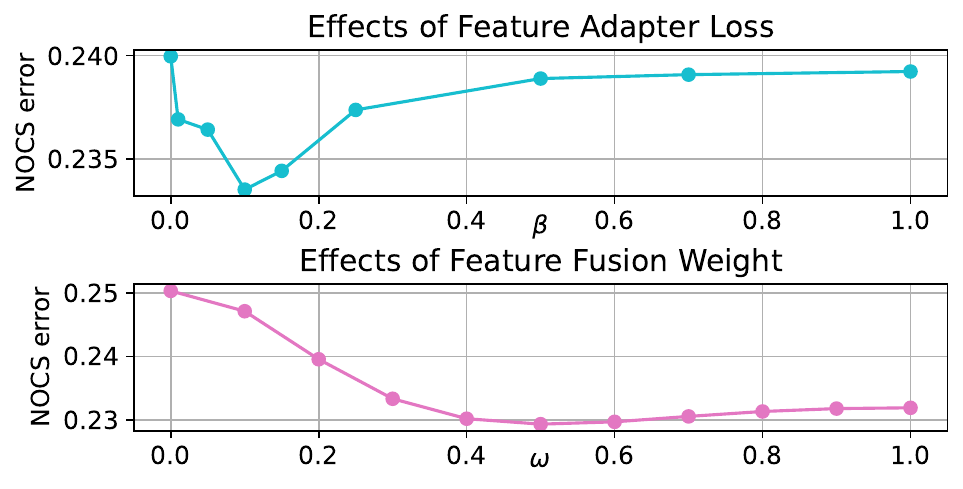}}
\vspace{-10pt}
\caption{Ablation studies on feature adapter losses (upper) and feature fusion weights (below).}
\label{fig:ablation1}
\vspace{-10pt}
\end{figure}

\noindent\textbf{Feature adapter loss study.}
The impact of $\beta$ for balancing feature adapter loss in Eq~\ref{eq:reconloss} and Eq~\ref{eq:tripletloss} is shown in Fig.~\ref{fig:ablation1}, where a weight of 0.1 yields the best NOC error.

\noindent\textbf{Feature fusion.}
We test $\omega$ for fusing DINOv2 with our geometry-aware features and find that $\omega=0.5$ minimizes NOC error (Fig.~\ref{fig:ablation1}), balancing both feature types.


\begin{table}[]

\centering
\setlength{\tabcolsep}{5pt}
\resizebox{235pt}{!}{%

\begin{tabular}{ccc|cc|ccccccc}

\toprule
\multicolumn{3}{c|}{\textbf{Dense alignment losses}}          &  \multicolumn{2}{c|}{\textbf{9 Categories}}  & \multicolumn{7}{c}{\textbf{Accuracy per parameter $\uparrow$}} \\ 
+$\mathcal{L}_{\text{NOC-A}}$       & +$\mathcal{L}_{\text{Mask}}$       & +$\mathcal{L}_{\text{Depth}}$       & \textbf{Cat.} $\uparrow$&\textbf{Inst.} $\uparrow$  && \textbf{Tr} $\uparrow$ & & \textbf{Sc} $\uparrow$  && \textbf{Ro}$\uparrow$&    \\\midrule
  -&-&- &  19.31 &  23.93   && 44.89   && 44.05 && 48.18    \\
    -   &  -           & \checkmark   & 20.47    &  24.87  && 45.18 && 46.92  && 49.72   \\
          -& \checkmark       & -   &  15.01    &   18.92     && 36.95 && 48.33  && 36.62     \\
\checkmark&-&-         &  18.45    &   23.76  && 41.46  &&  48.63 &&   48.90    \\
-& \checkmark& \checkmark     &  22.27   & 26.57    && 45.86 && 49.56  && 50.44   \\
\checkmark& \checkmark& - & 16.87     &  21.77    &&  39.69  &&  49.02  && 41.56  \\
\checkmark&-& \checkmark &  22.58   & 26.95   && 45.86  && 51.88  && 52.40  \\
\checkmark& \checkmark& \checkmark    & \textbf{23.60}     & \textbf{28.36}   && \textbf{46.67}  && \textbf{52.73} &&\textbf{52.49} \\
\bottomrule

\end{tabular}
    
}
\vspace{-10pt}
\caption{\textbf{Ablation study in dense image-based alignment loss}}
\vspace{-14pt}
\label{tab:ablation_refineloss}
\end{table} 

\noindent\textbf{Dense alignment loss study.}
Table~\ref{tab:ablation_refineloss} presents the impact of each loss term. $\mathcal{L}_{\text{NOC-A}}$ (Eq~\ref{eq:fine_noc}) improves scaling and rotation accuracy but degrades translation, while $\mathcal{L}_{\text{depth}}$ (Eq~\ref{eq:fine_depth}) enhances translation and rotation. Combining both losses improves all three parameters, and incorporating silhouette information from $\mathcal{L}_{\text{mask}}$ yields the best results.


\section{Conclusion}

We propose a zero-shot 9-DoF pose estimation method to align an inexact 3D model with its object in a single image. Our method enhances a foundation feature space to be geometry-aware, mitigating ambiguities from symmetrical parts, and leverages it to  enable initial pose estimation via simple nearest-neighbor matching. This space also facilitates pose refinement via dense alignment in a normalized coordinate space, which enhances robustness to texture mismatches and geometric differences between the input image and the 3D model.
Our method is fully self-supervised, requiring only a lightweight adapter trained on a small number of ScanNet25K categories, with no scene-level pose annotations.
On nine seen categories from ScanNet25K, it trails the state-of-the-art supervised SPARC $(-3.8\%)$ but outperforms weakly supervised competitors and the supervised ROCA $(+4.2\%, +2.7\%)$. However, on the newly introduced SUN2CAD dataset with 20 unseen categories, it achieves state-of-the-art generalization with a large margin $(+12.7\%)$, demonstrating its potential for scalable, category-agnostic alignment. 
{
    \small
    \bibliographystyle{ieeenat_fullname}
    \bibliography{bibb}
}
\clearpage
\setcounter{page}{1}
\vspace{0.1pt}
\maketitlesupplementary

\section*{Appendix}
In this Appendix, we provide additional clarifications, experiments, and results as follows:
\begin{itemize}
    \item Appendix~\ref{supp:ourimplementation}: Additional implementation details of our method
    \item Appendix~\ref{supp:augmentation}: Implementation details of rendered image augmentation
    \item Appendix~\ref{supp:competitors}: Implementation details of competitors and ablation choices
    \item Appendix~\ref{supp:new_dataset}: Additional details on SUN2CAD dataset
    \item Appendix~\ref{supp:moreexp}: Additional experimental results
    \item Appendix~\ref{supp:qual}: Additional qualitative results
    \item Appendix~\ref{supp:ablation}: Additional ablation studies
    \item Appendix~\ref{supp:failurecase}: Failure cases
    \item Appendix~\ref{supp:societal}: Societal impact
\end{itemize}

\section{Additional Implementation Details of Our method}
\label{supp:ourimplementation}

\subsection{Training details and hyperparameters}
For training the feature adapter, we use the AdamW optimizer~\cite{adamw} with a constant learning rate of $3e^{-4}$,
a batch size of 140, and
$\tau^{+}_{\text{dist}}$ = 0.02, $\tau^{-}_{\text{dist}}$ = 0.4, $\tau^{-}_{\text{feat}}$ = 0.75, $\alpha$ = 0.5, $\beta$ = 0.1, $\omega$ = 0.5.
The dense alignment is optimized with the Adam optimizer at a constant learning rate of 0.005 and 
$\lambda_\text{NOC-A}$ = 0.33, $\lambda_\text{m}$ = 3.0, $\lambda_\text{d}$ = 0.27. 
\subsection{Template Rendering}
\label{supp:template}

\begin{figure}[]
\centerline{\includegraphics[width=236pt]{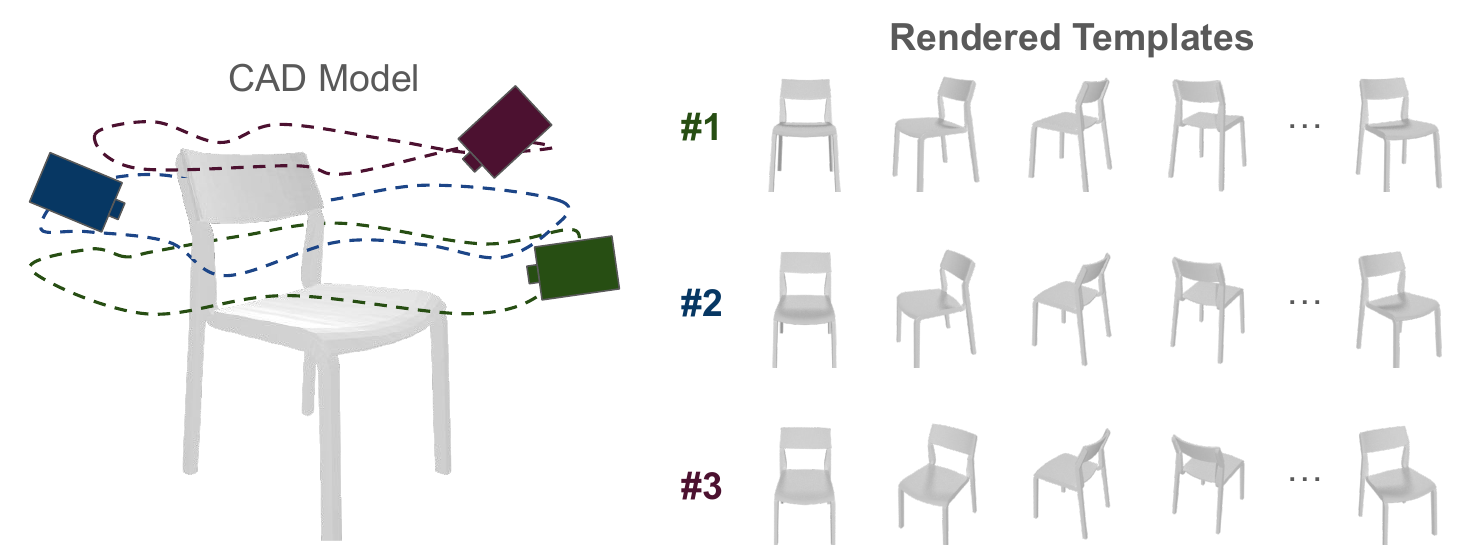}}
\caption{\textbf{Rendered templates used in a feature voxel grid and the geometry-aware feature adapter. } }
\label{fig:rendered}
\end{figure}

To render templates that cover the entire CAD model, we render each CAD model into 36 templates using Blender, varying 3 elevation angles and 12 azimuth angles. Each angle is randomly sampled from a mean of $[10, 20, 30]^\circ$ for elevation and $[0, 30, 60, \dots, 330]^\circ$ for azimuth, with a standard deviation of 2, allowing us to capture a broader range of perspectives around the CAD model.
The examples of templates are shown in Figure~\ref{fig:rendered}.

\subsection{Feature Voxel Grid}
We used 36 rendered templates with corresponding NOC pairs, augmenting each template into 7 variations (see Appendix~\ref{supp:augmentation}), resulting in a total of 288 images (36 original + 252 augmented) for constructing a feature voxel grid of size $100\times 100 \times 100$. Augmented image features were averaged with a weight of 0.071 per image, while rendered images were assigned a weight of 0.5.
To smooth the voxel grid, we applied a two-step downsampling and upsampling process using linear interpolation. The grid was first downsampled by factors of 2 and 4, then upsampled back to its original size. The final voxel grid was obtained by a weighted average of these versions, with weights of 0.6 for the original size, 0.25 for the 2× downsampled version, and 0.15 for the 4× downsampled version.

\subsection{Feature Adapter Training Data}

To generate our training data, we use ShapeNet~\cite{shapenet} dataset, which contains a collection of normalized, canonically aligned 3D models. We select models from nine categories that also appear in ScanNet25k~\cite{scannet} (bathtub, bed, bin, bookshelf, cabinet, chair, display, sofa, and table), totaling 2k 3D models. Each model is rendered into 36 templates, following Section~\ref{supp:template}, resulting in  72k rendered training images and NOC pairs.  Each template image is further augmented into seven variations and filtered (see Section~\ref{supp:augmentation}), leading to a final dataset of 300k training images for the feature adapter.
Note that we use this feature adapter, which is trained on only nine object categories, to evaluate 20 unseen categories in SUN2CAD.

To validate the design choices, we reserved 2k annotations from the ScanNet training set and used them for validation purposes only.

\subsection{Object Mask Generation}
For processing ScanNet25k~\cite{scannet} scene images, we use object bounding boxes and retrieved CAD models from ROCA~\cite{roca}. We then apply Segment Anything (SAM) with ViT-G~\cite{vit} to generate segmentation masks from the bounding box prompts.

\subsection{Metric Depth Estimator}
We fine-tuned metric depth estimators on the training images of ScanNet25k~\cite{scannet} and SUN-RGBD~\cite{sunrgbd} in the SUN2CAD dataset for use in comparisons in Section~\ref{sec:expresults}. We follow the codebase of DepthAnything~\cite{depthanything} by fine-tuning their pre-trained ViT-L relative depth estimator with image-metric depth map pairs from each dataset.
The training was performed with a learning rate of $5e^{-5}$ and a batch size of 24.

For ScanNet25k, we used 20k image-depth pairs from the training set, splitting them into 19k for training and 1k for validation. Depth maps were masked to retain only pixels with values $>0.01$, and the mask was dilated by 11 pixels to reduce aliasing and noise.

For SUN2CAD, we used 7k images from SUN-RGBD that are in the scene and do not overlap with scenes that appeared in our 550 testing images. We split them into 6.5k for training and 500 for validation. Since SUN-RGBD contains images from multiple sources and cameras, we first normalized all depth maps by resizing them to $518\times392$ and converting them into canonical inverse depth maps using $C = f/D$ (following~\cite{depthpro}), where $f$ is the focal length, $D$ is the metric depth, and $C$ is the inverse depth. We applied the same depth masking and dilation procedure as in ScanNet25k.
We then fine-tuned DepthAnything on the masked inverse depth maps. During inference, metric depth maps were reconstructed using the focal length of each test image via $D = f/C$.

In Section~\ref{sec:diffcadsplit}, we use a metric depth estimator that has never been trained on ScanNet~\cite{scannet} dataset for a fair comparison with DiffCAD~\cite{diffcad}, which is the official DepthAnything~\cite{depthanything} Indoor Metric Depth Estimator (ViT-L) trained on NYU dataset~\cite{nyu}.

\subsection{Visualization of Learned Space}
\label{sec:fig_explain}
Here we provide implementation details for the feature visualization in Fig.~\ref{fig:latent} in the main paper. Given a rendered image (Row 1 or 3), we extract a patchwise feature map using either DINOv2 or our feature adapter and obtain its corresponding ground-truth NOC map rom the CAD model we rendered. We then reduce the feature dimensions to two using PCA and visualize them as 2D points in the last two columns. Each feature point is colored based on its corresponding NOC value by mapping (x, y, z) into (r, g, b). The same procedure is applied to real image inputs (Row 2 or 4) using the ground-truth, pose-aligned CAD model to generate the NOC maps.


\section{Implementation Details of Rendered Image Augmentation}
\label{supp:augmentation}
\label{supp:genimg}

Renderings often have limited texture and solid color backgrounds, creating a domain gap with real images. We follow DST3D~\cite{dst3d} to generate realistic renderings with natural backgrounds, using Stable Diffusion (SD)~\cite{stablediffusion}. While DST3D uses ControlNet~\cite{controlnet} to guide image generation with a Canny edge map, we further enhance image fidelity by incorporating multiple visual prompts (Canny edge, depth, and sketch of the renderings) through UniControlNet~\cite{unicontrolnet}.

\begin{figure}[]
\centerline{\includegraphics[width=236pt]{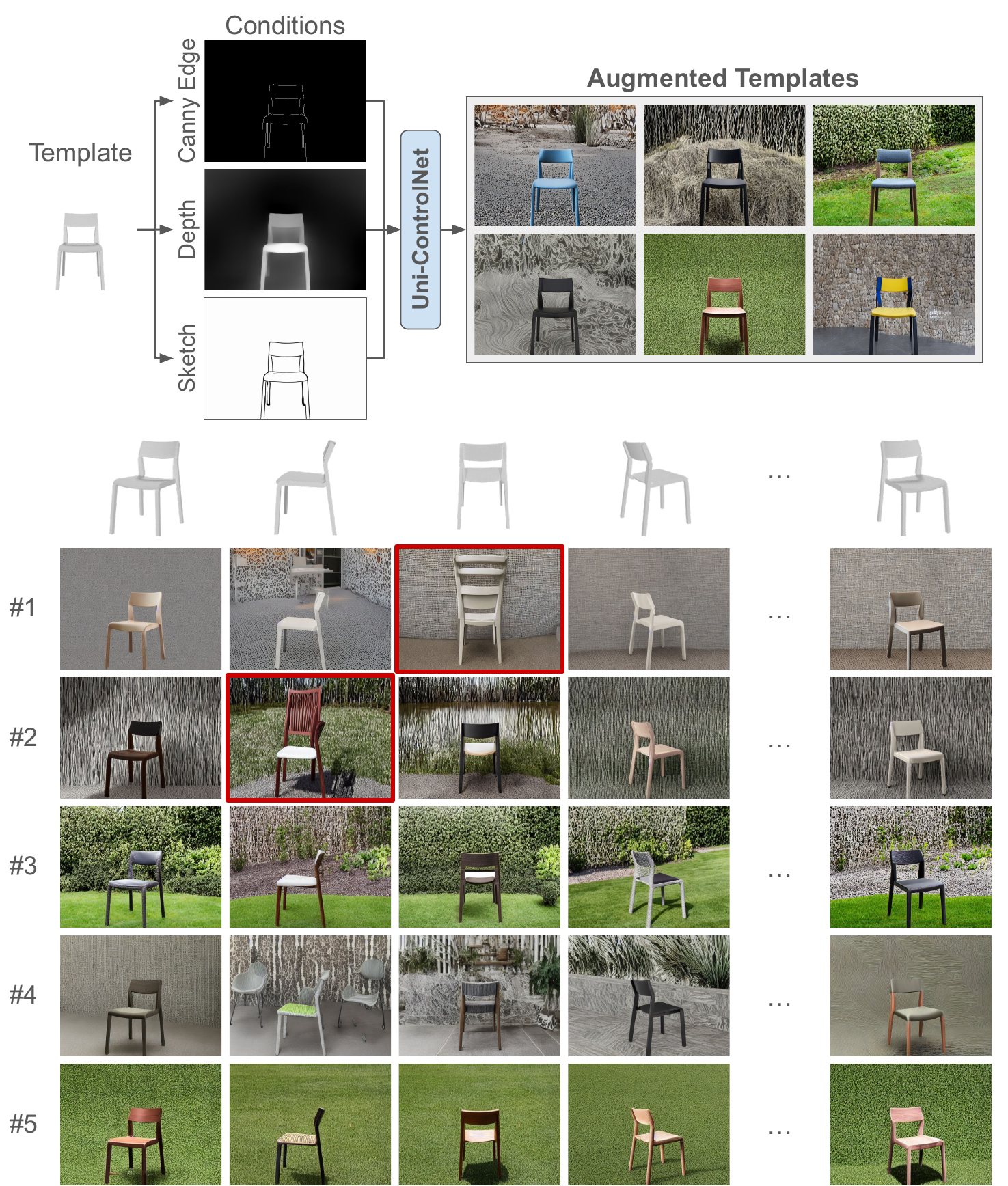}}
\caption{\textbf{Augmented templates used in a feature voxel grid and the geometry-aware feature adapter.} Annotated red boxes refer to invalid generated images with a wrong viewpoint shifted from their original renderings on the top row.}
\label{fig:augment}
\end{figure}

However, rarely seen images, such as the back of a cabinet, are challenging for the diffusion model to generate. Such generated images result in artifacts that reduce pose prediction performance. To address this, we propose filtering out poorly generated images based on our NOCS prediction error (See an ablation study in Appendix~\ref{supp:ab_traindata}). Examples of augmented templates are shown in Figure~\ref{fig:augment}, with filtered-out templates highlighted in red boxes.

\section{Implementation Details of Competitors and Ablation Choices}
\label{supp:competitors}
In this section, we provide implementation details for evaluating DiffCAD, FoundationPose, and SPARC in our setting, where scores are computed using their official code. For ROCA, scores are reported as provided in their paper. We also provide details for the dense alignment using features (FM), as presented in the ablation study in Section~\ref{sec:ablation_pipeline}.


\subsection{DiffCAD}
We use DiffCAD's~\cite{diffcad} official model weights and code to generate eight pose hypotheses per sample across six categories in their ScanNet25k test subset. We then report the best hypothesis's alignment accuracy in Table~\ref{tab:newmaintable}, denoted as DiffCAD (GT), as a reference to validate our reproduction of their results.

To compute DiffCAD (mean), we calculate translation, scale, and rotation errors for each hypothesis, average them per parameter, and determine alignment accuracy based on standard thresholds used in previous works~\cite{roca} as:
translation error $\leq 20$ cm, rotation error $\leq 20^\circ$, and scale error $\leq 20\%$ relative to the ground truth.

For DiffCAD (err), we use their codebase to select the best hypothesis based on transformation error on their predicted 2D-3D correspondences. Given $X$ as the set of selected 2D coordinates with depth values, $Y_i$ as the predicted 3D coordinates for the $i$th hypothesis, and $T_i$ as the solved transformation, the optimal hypothesis is the one minimizing $||Y_i \cdot T_i^T - X||_2$.

\subsection{FoundationPose}
We use FoundationPose~\cite{foundationpose}'s officially released model to generate 6-DoF poses on ScanNet25k~\cite{scannet} and SUN2CAD datasets. 

For ScanNet25k, we obtain a pair of a 2D bounding box and its corresponding CAD model from ROCA's results. These bounding boxes are processed into 2D object masks using SAM~\cite{sam}, consistent with our method. We also use the same fine-tuned DepthAnything~\cite{depthanything} to predict a depth map for each image. To enable fair comparison with our 9-DoF method, we scale the CAD models in two ways: 1) by ground truth scale for the 6-DoF setting and 2) by ROCA's predicted scale for the 9-DoF setting.

For the SUN2CAD dataset, we use a pair of 2D masks and CAD models as input, with depth maps generated by fine-tuned DepthAnything on the SUN RGB-D training split. As with ScanNet25k, we scale the CAD models by their ground-truth scale for the 6-DoF setting.

\subsection{SPARC}
We follow SPARC~\cite{sparc}'s officially released model weights and code to produce 9-DoF alignment prediction of the SUN2CAD dataset.
We utilize depth maps from fine-tuned DepthAnything~\cite{depthanything} on SUN RGB-D~\cite{sunrgbd}, along with 2D bounding boxes and CAD model pairs from the dataset, as used by other competitors.

In addition to RGB-D images and detected objects, SPARC requires surface normals for both the RGB images and CAD model point clouds. We sample 1,000 surface points and their corresponding normals from each CAD mesh using Trimesh. For image normals, we use Metric3Dv2~\cite{metric3dv2}, a state-of-the-art zero-shot normal estimator, to generate them for all images.

We follow the inference procedure instructed in the paper by initializing the translation using the $x$ and $y$ position at the middle point of a 2D bounding box and $z=3$. We apply median scaling of the 3D model's category and vary azimuth rotation across $[0, 90, 180, 270]^\circ$ with a fixed elevation angle of $15^\circ$, resulting in 4 hypotheses. Each hypothesis is refined 3 times, and the best hypothesis is selected using their trained classifier for further pose evaluation.


\subsection{Feature-based Dense Alignment}
This section provides detailed information on dense alignment using DINOv2 features for ablation studies in Section~\ref{sec:ablation_pipeline}. The method is inspired by the \emph{featuremetric optimization} refinement method from FoundPose~\cite{foundpose}. 
The goal of featuremetric optimization (FM) is to refine initial 2D-3D correspondences by minimizing projection errors using foundation features. 
This addresses the misalignment of large $14 \times 14$ patch-wise features, whose derived 2D coordinates (e.g., the patch center) may not precisely correspond to their 3D counterparts.

To mitigate this, the method optimizes pose by minimizing discrepancies between features at 3D coordinates (from a CAD model's rendered templates) and their corresponding 2D projections in the input image’s feature map. The objective function is:

\begin{equation}
\sum_{(x_i,p_i) \in T_t}  \rho \left( p_i-F_q \left( \pi(x_i)\right)\right),
\end{equation}

where $\rho$ is a cost function, $(p_i,x_i) \in T_t$ is a 3D coordinates and its paired feature, $F_q$ is a 2D feature map, and $\pi$ is a learnable 2D projection function.
Following the original algorithm, we reimplement the method using our feature voxel grid to represent 3D model features. We initialize $\pi$ with the coarse alignment pose and tune other hyperparameters for optimal performance.
We employ PyTorch3D~\cite{pytorch3d} and optimize the loss using Adam with a learning rate of 0.005. We use L1 loss as the cost function and bilinear interpolation for 2D feature sampling via grid sampling.


\section{Additional Details on SUN2CAD Dataset}
\label{supp:new_dataset}
To evaluate our alignment method on unseen or less common object categories, we establish a new inexact match 9-DoF pose alignment test set spanning 20 categories with 550 images.

The primary challenge in aligning images with 3D models is the ambiguity in z-axis translation and scaling, which persists even with manual alignment. We address this using 3D bounding box annotations from the SUN-RGBD dataset~\cite{sunrgbd}, which provides real RGB-D images of room scenes derived from 3D scans.
These bounding boxes were used as our initial pose, as the bounding box centroid will be a translation, the bounding box size will be used as a scale, and the annotated bounding box orientation will be treated as a rotation. 

To map between the 3D model and 2D image, we select LVIS categories~\cite{lvis} that are contained in both SUN-RGBD and CAD models from Objaverse~\cite{objaverse} or ShapeNet~\cite{shapenet}. For each SUN-RGBD object, we manually choose the most suitable CAD model of the same category and render it to fit the 3D bounding box, then refine its pose by manually rotating the 3D model for more precise alignment. To minimize annotation errors, we exclude objects with incorrect 3D bounding boxes or insufficient edge cues for alignment.
We also label object symmetry types following the ScanNet25K baseline: asymmetry, 2-side symmetrical, 4-side symmetrical, and all-side symmetrical.

LVIS category selection is prioritized based on 2D object-CAD similarity, frequency in SUN-RGBD scenes, and object categories with varied and distinct shapes or parts that are different from ScanNet25k's 9 original categories.
Initially considering over 50 categories, we finalize 20 categories: basket, bicycle, blender, broom, clock, coffee\_maker (coffmkr), crib, fire\_extinguisher (fireext), keyboard (keybrd), ladder, lamp, mug, piano, printer, remote\_control, shoe, telephone, toaster\_oven, vase, and water\_bottle. The quantity of each category is shown in Table~\ref{tab:sun2cad}.

Additionally, we aim to simulate a real-world scenario where a user selects a CAD model and object category, and the system automatically aligns it with the scene. Instead of relying on high-quality 2D segmentation masks provided by SUN-RGBD, we predict the segmentation using Grounded-SAM~\cite{groundedsam}, which enables automatic extraction of segmentation masks based on object category.

The samples of SUN2CAD dataset are shown in Figure~\ref{fig:sun2cad_ran_p1}, Figure~\ref{fig:sun2cad_ran_p2}, and Figure~\ref{fig:sun2cad_ran_p3}.

\section{Additional Experimentsal Results}
\label{supp:moreexp}


\subsection{Additional Results on Experiment~\ref{sec:mainexp}}
We present a detailed comparison of translation, rotation, and scaling accuracies across competitors on the ScanNet25k dataset in Table~\ref{tab:scannet_full}. All accuracies are computed using the same thresholds defined in Section~\ref{sec:expresults}.

In the weakly supervised 9-DoF setting, our method outperforms the 9D adapted FoundationPose~\cite{foundationpose}, designed for exact 6-DoF pose matching, in both translation and rotation for 7/9 categories, achieving better average parameter accuracies. When compared to supervised 9-DoF, ROCA~\cite{roca}, we achieve better translation accuracy in 8/9 categories and rotation accuracy in 4/9 categories. However, when compared to SPARC~\cite{sparc}, the strongest supervised baseline, our method exceeds performance in only 2/9 categories for both translation and rotation, revealing a remaining performance gap. Notably, the two supervised methods benefit from either learning object scaling during training or incorporating median object scaling during initialization, whereas our method does not rely on such priors.

In the weakly supervised 6-DoF setting, our adapted 6D solution surpasses FoundationPose~\cite{foundationpose} in translation for 7/9 categories and in rotation for 8/9 categories, also achieving higher mean accuracy for both parameters.

For a detailed comparison with DiffCAD~\cite{diffcad}, the state-of-the-art weakly supervised 9-DoF estimator, we present results in Table~\ref{tab:scannet_diff_supp}. Our method outperforms DiffCAD (mean) in translation and rotation for 5/6 categories and scaling for 4/6 categories. Additionally, we exceed DiffCAD (Err) in translation, scaling, and rotation for 4/6 categories. Overall, our approach achieves the highest average accuracy across all parameters.

\subsection{Additional Results on Experiment~\ref{sec:newdatasetexp}}
We also present a detailed comparison of each transformation parameter across competitors on the SUN2CAD dataset in Table~\ref{tab:sun2cad_tsr}. Alignment accuracies are computed using the same thresholds defined in Section~\ref{sec:expresults}. We also provided the mean and standard error statistics of each parameter for comparison.
In the 9-DoF setting, we beat DINOv2 in both accuracies and errors for all parameters. We also beat SPARC~\cite{sparc} in translation and scaling accuracy, but for rotation, we have a lower average rotation error by $-7.9\%$ with p-value $=0.0219$, which is significant.
These results indicate that SPARC struggles with the translation and scaling of unseen objects, while it remains somewhat effective in rotation but not as much as our method.
While in the inexact 6-DoF setting, we outperform FoundationPose~\cite{foundationpose} in all metrics.

\begin{table*}[htbp]
    \centering
    \resizebox{480pt}{!}{%
        \begin{tabular}{c|c|c|ccc|ccc|ccc|ccc|ccc}
            \toprule
            \textbf{Pose} & \textbf{Sup} &
            \textbf{Method}  &
            \multicolumn{3}{c|}{\textbf{bathtub}} & \multicolumn{3}{c|}{\textbf{bed}} & 
            \multicolumn{3}{c|}{\textbf{bin}} & \multicolumn{3}{c|}{\textbf{bkshlf}} &\multicolumn{3}{c}{\textbf{cabinet}} 
\\ 
            
           && &  Tr & Sc & Ro & Tr & Sc & Ro & Tr & Sc & Ro & Tr & Sc & Ro & Tr & Sc & Ro    \\
            \midrule
\multirow{4}{*}{9D}           &\cmark& ROCA~\cite{roca}  & \second{}48.9 & \first{}\textbf{77.8} & \second{}55.6 & 21.2 & \first{}\textbf{87.9} & \third{}51.5 & 56.4 & \first{}\textbf{76.0} & \first{}\textbf{57.3} & \third{}21.3 & \second{}46.2 & \second{}66.5 & \third{}19.3 & \second{}59.8 & \second{}64.2 \\ 
&\cmark&SPARC~\cite{sparc}  & \first{}\textbf{51.6} & \second{}69.6 & \first{}\textbf{68.5} & \first{}\textbf{40.0} & \second{}80.0 & \second{}56.9 & \second{}62.8 & \second{}71.3 & \second{}55.3 & \first{}\textbf{22.7} & \first{}\textbf{56.6} & \first{}\textbf{68.7} & \first{}\textbf{27.1} & \first{}\textbf{64.5} & \first{}\textbf{68.9} \\
&$\star$&FoundationPose~\cite{foundationpose} (for 9D) & 35.1 & \first{}\textbf{77.8} & 37.4 & \second{}29.2 & \first{}\textbf{87.9} & 36.9 & \third{}61.5 & \first{}\textbf{76.0} & \third{}38.5 & 6.5 & \second{}46.2 & 5.5 & 6.8 & \second{}59.8 & 7.2 \\ 
&$\star$&\textbf{Ours}  & \third{}43.8 & 51.7 & \third{}39.3 & \second{}29.2 & 46.2 & \first{}\textbf{61.5} & \first{}\textbf{67.3} & 70.8 & 34.6 & \second{}22.3 & 38.3 & \third{}52.3 & \second{}20.2 & 39.7 & \third{}55.1  \\ \midrule

\multirow{2}{*}{6D}&$\star$ &FoundationPose~\cite{foundationpose}  & \first{}\textbf{51.1} & - & 37.8 & 27.7 &- & 30.8 &64.7 &- & 30.8 & 13.9 &- &12.4 & 8.0 &- & 10.4  \\
&$\star$&\textbf{Ours} (for 6D) & 48.9 & - & \first{}\textbf{38.6} & \first{}\textbf{29.7} & - &\first{}\textbf{59.4} & \first{}\textbf{69.8} &- & \first{}\textbf{40.1} &\first{}\textbf{23.6} &- & \first{}\textbf{46.7} & \first{}\textbf{26.3} & - & \first{}\textbf{49.0}\\
           \midrule \toprule
                     \textbf{Pose} & \textbf{Sup} &      \textbf{Method}          & \multicolumn{3}{c|}{\textbf{chair}} & 
            \multicolumn{3}{c|}{\textbf{display}} & \multicolumn{3}{c|}{\textbf{sofa}} & 
            \multicolumn{3}{c|}{\textbf{table}} & 
            \multicolumn{3}{c}{\textbf{Avg.}} \\
           && &  Tr & Sc & Ro & Tr & Sc & Ro & Tr & Sc & Ro & Tr & Sc & Ro & Tr & Sc & Ro \\ \midrule
           
\multirow{4}{*}{9D}           &\cmark& ROCA~\cite{roca}  & 45.2 & \first{}\textbf{93.6} & \third{}57.9 & \third{}43.8 & \first{}\textbf{62.2} & \second{}51.6 & \third{}27.8 & \third{}59.8 & \third{}63.9 & 21.5 & \second{}69.0 & \first{}\textbf{61.8} & \third{}36.0 & \second{}76.8 & \second{}59.5 \\ 
           &\cmark&SPARC~\cite{sparc}  & \first{}\textbf{57.7} & 93.2 & \first{}\textbf{71.5} & \first{}\textbf{48.6} & \second{}54.7 & \third{}45.8 & \second{}40.2 & \second{}75.2 & \first{}\textbf{79.4} & \first{}\textbf{27.5} &\first{}\textbf{74.6} & \second{}59.5 & \first{}\textbf{44.8} & \first{}\textbf{78.3} & \first{}\textbf{65.6} \\
           & $\star$ &FoundationPose~\cite{foundationpose} (for 9D)   & \third{}46.5 & \first{}\textbf{93.6} & 48.3 & 34.4 & \first{}\textbf{62.2} & 36.1 & 22.7 & \third{}59.8 & 36.1 & \second{}26.9 & \second{}69.0 & 42.3 & 35.1 & \second{}76.8 & 37.4  \\
&$\star$&\textbf{Ours}  & \second{}57.3 & 76.7 & \second{}61.9 & \second{}44.3 & 40.9 & \first{}\textbf{53.4} & \first{}\textbf{47.4} & \first{}\textbf{77.9} & \second{}73.7 & \third{}24.3 & 47.4 & \third{}44.8 & \second{}43.2 & 60.3 & \third{}54.2\\ \midrule

\multirow{2}{*}{6D}&$\star$&FoundationPose~\cite{foundationpose} &48.4 &- & 50.6 & 52.5 & -& 40.2 &37.1 &- & 44.3 & \first{}\textbf{32.8} &- & \first{}\textbf{40.9} & 39.7 & - & 40.2   \\
&$\star$&\textbf{Ours} (for 6D) & \first{}\textbf{61.3} &- &\first{}\textbf{65.6} & \first{}\textbf{67.2} &- &\first{}\textbf{68.4} & \first{}\textbf{54.6} &- & \first{}\textbf{77.4} & 27.7 & - & 38.3 & \first{}\textbf{48.2} & - & \first{}\textbf{54.8} \\
            
            \bottomrule
        \end{tabular}
    }
    \caption{\textbf{Detailed comparison on ScanNet25k~\cite{scannet} across 9 object categories.} Tr, Sc, and Ro represent translation, scaling, and rotation alignment accuracy. Supervised and weakly supervised baselines are marked with `\cmark' and `$\star$' respectively.}
    \label{tab:scannet_full}
\end{table*}

\begin{table*}[htbp]
    \centering
    \resizebox{490pt}{!}{%
        \begin{tabular}{c|ccc|ccc|ccc|ccc|ccc|ccc|ccc}
            \toprule

            \textbf{Method}  &
            \multicolumn{3}{c|}{\textbf{bed}} & \multicolumn{3}{c|}{\textbf{bkshlf}} & 
            \multicolumn{3}{c|}{\textbf{cabinet}} & \multicolumn{3}{c|}{\textbf{chair}} &\multicolumn{3}{c|}{\textbf{sofa}} 
            &\multicolumn{3}{c|}{\textbf{table}} 
            &\multicolumn{3}{c}{\textbf{Avg.}} 
\\ 
            
 &  Tr & Sc & Ro & Tr & Sc & Ro & Tr & Sc & Ro & Tr & Sc & Ro & Tr & Sc & Ro  & Tr & Sc & Ro& Tr & Sc & Ro \\\midrule
DiffCAD (Mean) & 5.9 &35.9 & 43.8 & 6.0 & 10.8 & 69.6  &11.8 & 35.4 &54.9 & 38.5 & 51.0 & 68.5 & 11.5 & 34.5 & 78.9 & 8.7 & 12.4 & 39.5 & 20.2 & 33.4 & 59.0\\
  DiffCAD (Err) & \first{}\textbf{11.1} & \first{}\textbf{39.8}  & 56.9 &  \first{}\textbf{22.5} & 25.8 &\first{}\textbf{80.1} & 19.2  & \first{}\textbf{36.7} & 78.5 & 47.9 & 54.4& 74.5 & 21.8 &46.7 &82.3  & 13.9 & 23.9&\first{}\textbf{56.9} &28.6&40.5&70.6   \\
    \textbf{Ours} & 6.7 & 31.9 & \first{}\textbf{58.5} & 13.5 & \first{}\textbf{35.3} & 67.3 & \first{}\textbf{22.7} & 31.5 & \first{}\textbf{85.2} & \first{}\textbf{54.5} & \first{}\first{}\textbf{79.6} & \first{}\first{}\textbf{80.3} & \first{}\textbf{27.0} & \first{}\textbf{72.5} & \first{}\textbf{91.5} & \first{}\textbf{16.4} & \first{}\textbf{37.9} & 49.6 & \first{}\textbf{32.3} & \first{}\textbf{56.6} & \first{}\textbf{71.5}\\ \bottomrule

\end{tabular}

}
    \caption{\textbf{Detailed comparison on DiffCAD's ScanNet25k split across 6 object categories.} Tr, Sc, and Ro represent translation, scaling, and rotation alignment accuracy.}
    \label{tab:scannet_diff_supp}

\end{table*}

\begin{table*}[htbp]
\small
    \centering
        \begin{tabular}
        {c|c|c|ccc|ccc}
        \toprule
            \textbf{Pose} & \textbf{Sup} &
            \textbf{Method}  & \multicolumn{3}{c}{\textbf{Avg Acc.} $\uparrow$} & \multicolumn{3}{c}{\textbf{Avg Err. $\pm$ SE $\downarrow$} }  \\ 
             &  &
& Tr & Sc & Ro & Tr & Sc & Ro \\ \midrule
    \multirow{3}{*}{9D}  & \cmark & SPARC~\cite{sparc} & 27.82 & 31.53 & \first{}\textbf{42.00} & 0.4381 $\pm$ 0.0179 & 61.26 $\pm$ 2.79 & 53.12 $\pm$ 2.37\\
& - & DINOv2 & 55.39 & 32.17 &24.31 & 0.2698 $\pm$ 0.0152 & 68.65 $\pm$  5.70 & 62.88 $\pm$ 2.08\\
& $\star$ & \textbf{Ours} & \first{}\textbf{62.36} & \first{}\textbf{43.09} & 41.09 & \first{}\textbf{0.2374 $\pm$ 0.0134} & \first{}\textbf{41.17 $\pm$ 2.27} & \first{}\textbf{46.03} $\pm$ 1.98\\  \midrule
\multirow{2}{*}{6D}  & $\star$ & FoundationPose~\cite{foundationpose} & 55.81 &- &26.72 & 0.2778 $\pm$ 0.0174 & - & 94.88 $\pm$ 2.88\\
& $\star$ & \textbf{Ours} (for 6D)& \first{}\textbf{64.72} & - & \first{}\textbf{41.63} & \first{}\textbf{0.2235 $\pm$ 0.0134} &- & \first{}\textbf{44.61 $\pm$ 1.97}\\
            \bottomrule
        \end{tabular}
\normalsize
    \caption{\textbf{Comparison in single-view accuracy on SUN2CAD dataset.} Tr, Sc, and Ro represent translation, scaling, and rotation alignment accuracy/errors. Supervised, weakly supervised, and unsupervised baselines are marked with `\cmark', `$\star$', and `-' respectively.}
    \label{tab:sun2cad_tsr}
\end{table*}

\subsection{Comparison on Unseen Categories in ScanNet25k}

We removed one of the nine categories in ScanNet25k~\cite{scannet} used for training and tested single-view accuracies~\cite{diffcad} on this removed category to assess the robustness of unseen categories. 
To do a comparison, we introduce a new baseline called \emph{Real-data NOC regressor (NOC-R)}.
NOC-R is similar to NOC-S, except it is trained on real indoor scenes from ScanNet25k~\cite{scan2cad} dataset, with ground-truth NOC maps generated from CAD models and their pose annotations. This baseline represents what happens if a category-specialized model faces unseen categories.

\begin{figure*}[!htbp]
\centering
\includegraphics[width=495pt]{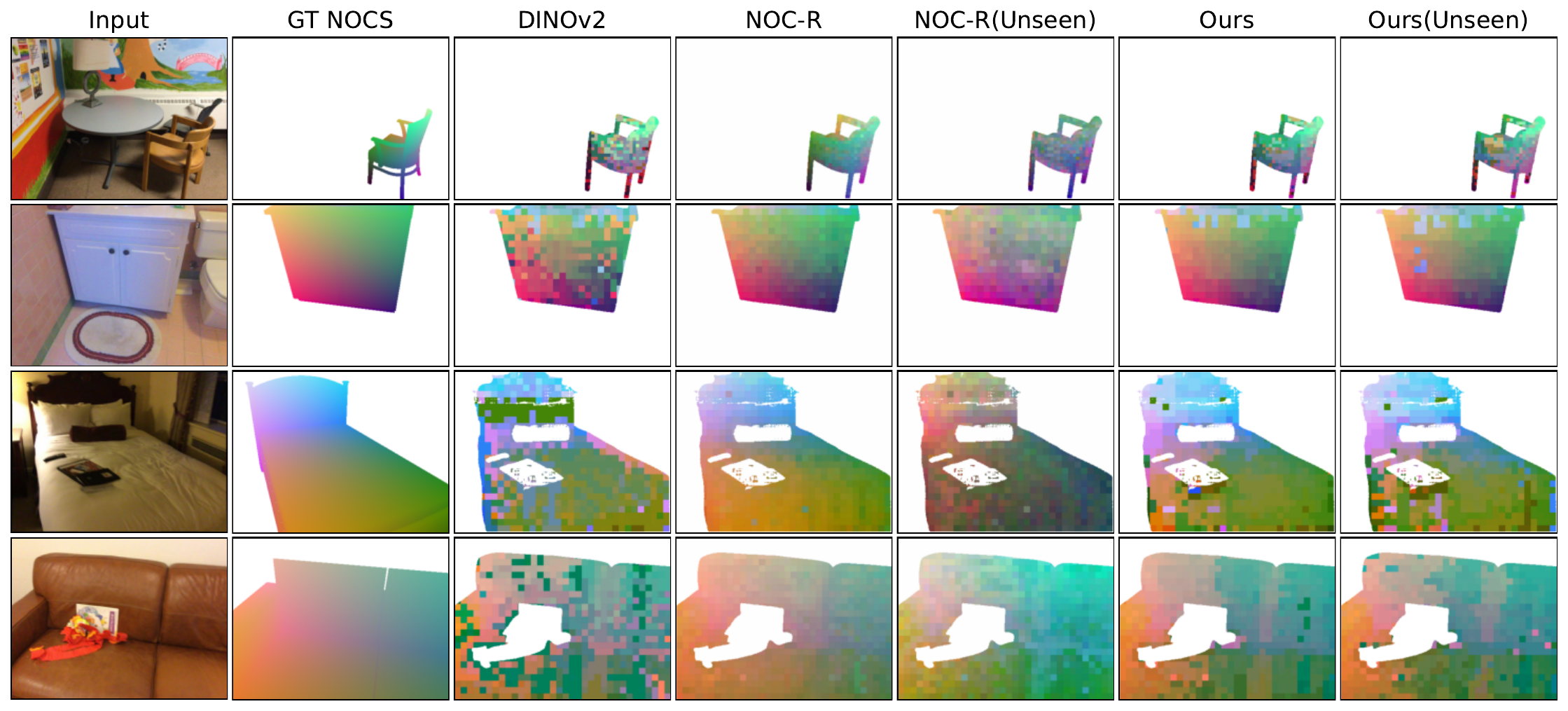}
\caption{\textbf{Qualitative results in NOC map prediction on unseen ScanNet25k categories.} Our method reliably generates more accurate NOC maps than DINOv2 and is robust against unseen categories, unlike NOC-R, which excels in standard settings but struggles in unfamiliar categories. }
\label{fig:nocsresults}
\end{figure*}

Table~\ref{tab:unseenclass} shows that our method experiences only a small drop of $\{-0.4\%,-0.8\%\}$ compared to our variant that trained on all categories, and it still outperforms DINOv2~\cite{dinov2} by $\{+5.6,+5.0\%\}$. In contrast, the strong supervised baseline NOC-R suffers a significant drop by $\{-19.2\%, -24.2\%\}$. Figure~\ref{fig:nocsresults} visualizes the differences in NOC map predictions for seen vs. unseen categories, highlighting wrong predicted distributions in unseen NOC-R compared to our unseen model.

\begin{table*}[htbp]
\centering

\small

\begin{tabular}{c|c|c|ccccccccc|cc}
\toprule

\textbf{Seen} & \textbf{Method}      & \textbf{Sup}     & \textbf{bathtub} & \textbf{bed} & \textbf{bin} & \textbf{bkshlf} & \textbf{cabinet} & \textbf{chair} & \textbf{display} & \textbf{sofa} & \textbf{table} & \textbf{Cat.$\uparrow$} & \textbf{Inst.$\uparrow$}  \\ \midrule
\multirow{2}{*}{\cmark}&NOC-R   & \cmark  &  \textbf{38.8} & \textbf{22.3} & \textbf{33.0} & \textbf{18.7} & \textbf{40.4} & \textbf{46.5} & \textbf{32.1} & \textbf{46.8} & \textbf{24.6} & \textbf{33.7} & \textbf{37.6}   \\
&\textbf{Ours}  & $\star$   &  31.6 & 13.7& 18.5 & 13.2 & 20.5 & 40.0 &24.7 & 39.5 & 22.8 & 24.9 & 30.1\\\midrule
\multirow{3}{*}{$\times$}&NOC-R   & \cmark  & 18.5 & 0.0 & 7.7 & 10.0 & \textbf{28.3} & 12.9 & 12.4 & 31.9 & 8.4 & 14.5 & 13.4 \\
&DINOv2     & -     &  22.7 & 10.2 & 14.9 & 5.8 & 18.7 & 34.9 & 12.6 &34.8 & 15.9 & 18.9 & 24.3 \\ 
&\textbf{Ours}  &  $\star$      &  \textbf{28.6} & \textbf{14.2} & \textbf{21.7} & \textbf{12.9} &23.5 & \textbf{38.3} & \textbf{20.9} & \textbf{37.6} & \textbf{22.2} &\textbf{24.5} & \textbf{29.3}  \\
\bottomrule
\end{tabular}
\normalsize
\caption{\textbf{Comparison in single-view accuracy on unseen categories on ScanNet25k.} Supervised, weakly supervised, and unsupervised baselines are marked with `\cmark', `$\star$', and `-' respectively. Our method maintains accuracy on unseen classes while the supervised baseline NOC-R completely fails to adapt.}
\label{tab:unseenclass}
\end{table*}

\subsection{Comparison on the Zero-Shot Capability of Feature Adapter}
We evaluate the zero-shot capability of geometry-aware features on unseen categories in the SUN2CAD dataset. In addition to Table~\ref{tab:sun2cad} in the main paper, we report alignment results using only pure geometry-aware features without fusion (Ours ($\omega=1$)), to assess whether our adapted features can inherit the generalizability of DINOv2 without being explicitly fused. As shown in Table~\ref{tab:geoonly_sun}, Ours ($\omega=1$) outperforms DINOv2 by $\{+2.9\%, +7.0\%\}$, demonstrating that our standalone features generalize well to unseen categories and surpass DINOv2 in the inexact CAD model alignment task. Nevertheless, our proposed fused feature version (Ours) still achieves the best performance overall.
\begin{table*}[]
\addtolength{\tabcolsep}{-0.3em}
\resizebox{\textwidth}{!}{
\begin{tabular}{c|c|cccccccccccccccccccc|cc}
\toprule

 

  \textbf{Group} & \textbf{Method}  & \rotatebox[origin=c]{90}{\textbf{basket}} & \rotatebox[origin=c]{90}{\textbf{bicycle}} & \rotatebox[origin=c]{90}{\textbf{blender}} & \rotatebox[origin=c]{90}{\textbf{broom}} & \rotatebox[origin=c]{90}{\textbf{clock}} & \rotatebox[origin=c]{90}{\textbf{coffmkr}} & \rotatebox[origin=c]{90}{\textbf{crib}} & \rotatebox[origin=c]{90}{\textbf{fireext}} & \rotatebox[origin=c]{90}{\textbf{keybrd}} & \rotatebox[origin=c]{90}{\textbf{ladder}} & \rotatebox[origin=c]{90}{\textbf{lamp}} & \rotatebox[origin=c]{90}{\textbf{mug}} & \rotatebox[origin=c]{90}{\textbf{piano}} & \rotatebox[origin=c]{90}{\textbf{printer}} & \rotatebox[origin=c]{90}{\textbf{remote}} & \rotatebox[origin=c]{90}{\textbf{shoe}} & \rotatebox[origin=c]{90}{\textbf{phone}} & \rotatebox[origin=c]{90}{\textbf{oven}} & \rotatebox[origin=c]{90}{\textbf{vase}} & \rotatebox[origin=c]{90}{\textbf{bottle}} & \textbf{Cat.$\uparrow$} & \textbf{Inst.$\uparrow$}         \\ 
 
&& \#7&\#14&\#7&\#2&\#13&\#19&\#18&\#15&\#66&\#4&\#132&\#59&\#18&\#92&\#8&\#3&\#47&\#14&\#9&\#3&\#20&\#550\\
\midrule
\multirow{3}{*}{9D}& DINOv2
 & 0.0& \first{}\textbf{28.6} & \first{}\textbf{14.3} & \first{}\textbf{50.0} & \first{}\textbf{7.7} & 10.5 &0.0 & \first{}\textbf{13.3} &3.0 & 0.0 &5.4 &0.0 &44.4 &7.6 & 0.0 & \first{}\textbf{33.3} &6.4 & 7.1 & 0.0 & 0.0 & 11.6 & 7.3   \\  
 & Ours ($\omega$=1) &14.3 & 14.3 & 0.0 & \first{}\textbf{50.0} & \first{}\textbf{7.7} & \first{}\textbf{26.3} & 5.6 & 6.7 & 12.1 & 0.0 & \first{}\textbf{9.8} & \first{}\textbf{11.9} & \first{}\textbf{50.0} & 19.6 &0.0 &0.0 & \first{}\textbf{8.5} &42.9 & 11.1 & 0.0 & 14.5 & 14.3 \\
 & \textbf{Ours} &  \first{}\textbf{42.9}&21.4&\first{}\textbf{14.3}&\first{}\textbf{50.0}&\first{}\textbf{7.7}&21.1&\first{}\textbf{16.7}&6.7&\first{}\textbf{24.2}&\first{}\textbf{25.0}&6.1&10.2&\first{}\textbf{50.0}&\first{}\textbf{25.0}&\first{}\textbf{25.0}&\first{}\textbf{33.3}&\first{}\textbf{8.5}&\first{}\textbf{57.1} & 11.1 &\first{}\textbf{33.3}  &\first{}\textbf{24.5} &\first{}\textbf{17.6}   \\  

 \bottomrule
\end{tabular}
}
\caption{\textbf{Comparison in single-view accuracy on the zero-shot capability of feature choices on SUN2CAD dataset.} Our fused geometry-aware features achieve the highest accuracy. Even without fusion, pure geometry-aware features (Ours ($\omega$=1)) outperform DINOv2, demonstrating inherited generalization to unseen categories.
} 
\label{tab:geoonly_sun}
\end{table*}

\subsection{Comparison on Generalizability on Non-textured CAD Models}

We evaluate the texture sensitivity of our method by removing textures from textured CAD models and testing single-view alignment accuracies~\cite{diffcad} against models with textures. As shown in Table~\ref{tab:notexture}, our method shows a minimal drop of $\{-0.17\%,-0.05\%\}$ in mean accuracies, while DINOv2 got a significant decrease of $\{-0.98\%,-1.68\%\}$ mean accuracies. This represents our method's robustness to the textureless CAD model over vanilla foundation features.

\begin{table*}[]
\centering
\small

\begin{tabular}{c|c|ccccccccc|cc}
\toprule
 \textbf{Method} & \textbf{Texture}      & \textbf{bathtub} & \textbf{bed} & \textbf{bin} & \textbf{bkshlf} & \textbf{cabinet} & \textbf{chair} & \textbf{display} & \textbf{sofa} & \textbf{table} &  \textbf{ Cat.$\uparrow$} & \textbf{Inst.$\uparrow$} \\
\midrule
\multirow{2}{*}{DINOv2} & \checkmark   & 30.77 & 7.43 & 14.65 & 5.97 & 18.63 & 33.71 & 13.64 & 35.12 & 14.77 & 19.14 &24.13 \\
&$\times$  & 28.71 & 6.61 &14.58 &4.26 & 17.63 & 30.56 &11.30 & 35.44 & 14.40 & 18.16 &22.45    \\\midrule
\multirow{2}{*}{ \textbf{Ours} } &\checkmark  
& 35.89 & 10.74 & 19.89 & 14,48 & 21.46&38.53 &23.02 & 39.87 & 21.90 & 25.09 & 29.81\\ 
& $\times$  & 34.35 & 8.26 & 26.17 & 13.24 & 22.62 & 38.49 & 23.02 &36.07 & 22.04 & 24.92 &29.76   \\
                  
\bottomrule
\end{tabular}
\normalsize
\caption{\textbf{Comparison in single-view accuracy on removing CAD model texture on ScanNet25k.} Our method retains similar accuracies when removing a query CAD model's textures, compared to DINOv2, which shows more significant accuracy drops.}
\label{tab:notexture}
\end{table*} 

\subsection{Comparison on Robustness to Model Inexactness}
\begin{figure}[htbp]
\vspace{-10pt}

\centering
\includegraphics[width=\linewidth]{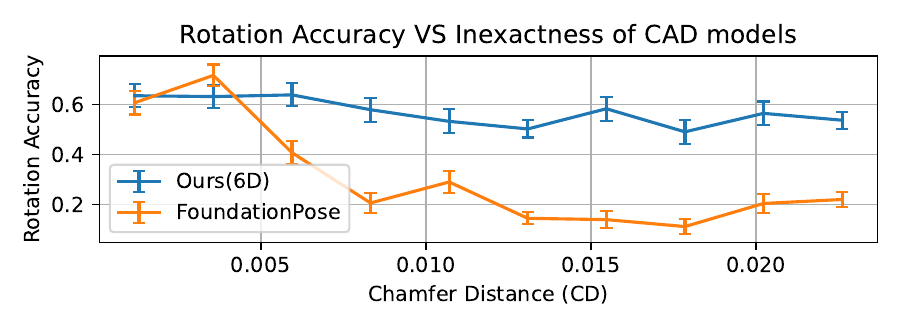}

\caption{
Rotation accuracy under varying CAD model inexactness, measured by Chamfer distance and averaged over 100 ScanNet25k images.
}
\label{fig:inexact}
\end{figure}

In Figure~\ref{fig:inexact}, we assess our method's sensitivity to input CAD model inexactness by replacing the ground-truth model with various models from the same category and reporting accuracy as a function of inexactness, measured by Chamfer distance. (We use rotation accuracy since their shared frontal alignment ensures the same optimal rotation, while shape differences affect translation and scale.) 
As inexactness increases, our method degrades far less than FoundationPose~\cite{foundationpose} (-0.14 vs -0.49), the only self-supervised SOTA testable here; DiffCAD~\cite{diffcad} predicts pose before CAD retrieval and cannot take an arbitrary model.

\subsection{Additional Comparison with FoundationPose}
To ensure a fair comparison with FoundationPose~\cite{foundationpose} under its original assumption used in its 6-DoF setting, we evaluate it using ground-truth depth maps (FoundationPose + GT Depth) and our previously introduced ground-truth object scaling.
Table~\ref{tab:found_scannet} shows a comparison of the ScanNet25k dataset. When given ground-truth depth maps, FoundationPose + GT depth achieves better performance than its default one (FoundationPose) with predicted depth maps for all categories.
 Our method, which used predicted depth, still surpasses their ground-truth depth version in 5/9 categories and in mean accuracies for both 9-DoF and 6-DoF settings by $\{+1.4\%,+1.7\%\}$. and $\{+2.6\%,+3.0\%\}$, respectively.
For a comparison in SUN2CAD dataset shown in Table~\ref{tab:tab:found_sun}, our adapted method surpasses FoundationPose with ground-truth depth 9/20 categories and tied in 3 categories. Although we still outperform them in mean accuracies by $\{+5.3\%,+1.2\%\}$.

\begin{table*}[htbp]
\centering
\resizebox{495pt}{!}{%

\begin{tabular}{c|c|c|ccccccccc|cc}
\toprule
\textbf{Group} &\multicolumn{2}{c|}{\textbf{Method}}     
 & \textbf{bathtub} & \textbf{bed} & \textbf{bin} & \textbf{bkshlf} & \textbf{cabinet} & \textbf{chair} & \textbf{display} & \textbf{sofa} & \textbf{table} & \textbf{Avg Cat.$\uparrow$} & \textbf{Avg Inst.$\uparrow$ }  \\ \midrule
\multirow{3}{*}{9D} & \multicolumn{2}{c|}{FoundationPose~\cite{foundationpose} (for 9D)} & 20.0 & 22.9  & 27.6  & 0.9 & 3.1  & 41.8  & 23.6  & 15.0  & 17.5  & 19.2  & 25.7  \\
&\multicolumn{2}{c|}{\cite{foundationpose} (for 9D) + GT depth} & \first{}\textbf{23.3} & \first{}\textbf{25.7} &\first{}\textbf{34.5} & 1.9 & 3.1 & 45.4 & 20.4 & 20.4 & \first{}\textbf{20.3} & 21.7 & 28.4  \\

 & \multicolumn{2}{c|}{\textbf{Ours}}&  16.7 & 18.6  & 22.8  & \first{}\textbf{12.7}  & \first{}\textbf{9.2}  & \first{}\textbf{49.3}  & \first{}\textbf{24.1} & \first{}\textbf{38.1}  & 16.5  & \first{}\textbf{23.1}  & \first{}\textbf{30.1}  \\ \midrule

\multirow{3}{*}{6D} & \multicolumn{2}{c|}{FoundationPose~\cite{foundationpose}}  & 22.5 & 21.4  & 37.5  & 6.1  & 5.8  & 44.5  & 30.4  & 29.2  & 27.1  & 24.9  & 31.1  \\
&\multicolumn{2}{c|}{\cite{foundationpose} + GT depth} &\first{}\textbf{25.8} & \first{}\textbf{27.1}  & \first{}\textbf{47.0}  & 9.0  & 7.3  & 46.7  & 32.5  & 38.1  & \first{}\textbf{33.1}  & 29.6  & 35.0  \\
 & \multicolumn{2}{c|}{\textbf{Ours} (for 6D)}&  20.8 & 25.7  & 27.6  & \first{}\textbf{19.8}  & \first{}\textbf{22.7}  & \first{}\textbf{56.1}  & \first{}\textbf{51.8}  & \first{}\textbf{45.1}  & 20.1  & \first{}\textbf{32.2}  & \first{}\textbf{38.0}  \\
 \bottomrule
\end{tabular}
}
\caption{\textbf{Comparison in NMS accuracy~\cite{vid2cad} on ScanNet25k~\cite{scan2cad} against FoundationPose~\cite{foundationpose}.} We additionally provide GT depth maps to FoundationPose to match their original setting. Our method still surpasses \cite{foundationpose} + GT depth in average alignment accuracies.
} 
\label{tab:found_scannet}
\end{table*}

\begin{table*}[]
\addtolength{\tabcolsep}{-0.3em}
\resizebox{\textwidth}{!}{
\begin{tabular}{c|c|cccccccccccccccccccc|cc}
\toprule

 

  \textbf{Group} & \textbf{Method}  & \rotatebox[origin=c]{90}{\textbf{basket}} & \rotatebox[origin=c]{90}{\textbf{bicycle}} & \rotatebox[origin=c]{90}{\textbf{blender}} & \rotatebox[origin=c]{90}{\textbf{broom}} & \rotatebox[origin=c]{90}{\textbf{clock}} & \rotatebox[origin=c]{90}{\textbf{coffmkr}} & \rotatebox[origin=c]{90}{\textbf{crib}} & \rotatebox[origin=c]{90}{\textbf{fireext}} & \rotatebox[origin=c]{90}{\textbf{keybrd}} & \rotatebox[origin=c]{90}{\textbf{ladder}} & \rotatebox[origin=c]{90}{\textbf{lamp}} & \rotatebox[origin=c]{90}{\textbf{mug}} & \rotatebox[origin=c]{90}{\textbf{piano}} & \rotatebox[origin=c]{90}{\textbf{printer}} & \rotatebox[origin=c]{90}{\textbf{remote}} & \rotatebox[origin=c]{90}{\textbf{shoe}} & \rotatebox[origin=c]{90}{\textbf{phone}} & \rotatebox[origin=c]{90}{\textbf{oven}} & \rotatebox[origin=c]{90}{\textbf{vase}} & \rotatebox[origin=c]{90}{\textbf{bottle}} & \textbf{Cat.$\uparrow$} & \textbf{Inst.$\uparrow$}         \\ 
 
&& \#7&\#14&\#7&\#2&\#13&\#19&\#18&\#15&\#66&\#4&\#132&\#59&\#18&\#92&\#8&\#3&\#47&\#14&\#9&\#3&\#20&\#550\\
\midrule
\multirow{3}{*}{6D} & FoundationPose~\cite{foundationpose} &  28.6 & \first{}\textbf{50.0} & \first{}\textbf{14.3} & 50.0 & 7.7 & 5.3 &11.1 & 33.3 & 33.3 &25.0 & 23.5 & 15.3 & 11.1 & 7.6 & \first{}\textbf{25.0} & 0.0 & 29.8 & 21.4 & 0.0 & 33.3 &21.3 & 20.4\\
  &\cite{foundationpose} + GT depth 
&28.6  &\first{}\textbf{50.0} & \first{}\textbf{14.3} & \first{}\textbf{100.0} & 7.7 & 10.5 & 11.1 & \first{}\textbf{53.3} & 31.8 & \first{}\textbf{75.0} & \first{}\textbf{47.7} & \first{}\textbf{27.1} & 16.7 & 9.8 & 0.0 & 0.0 &\first{}\textbf{31.9} & 0.0 & \first{}\textbf{55.6} & 33.3 & 30.2 & 29.5 \\

& \textbf{Ours} (for 6D) & \first{}\textbf{42.9} & \first{}\textbf{50.0}& \first{}\textbf{14.3} & 50.0 &\first{}\textbf{23.1} &\first{}\textbf{36.8} &\first{}\textbf{16.7} &20.0 & \first{}\textbf{39.4} & 50.0 & 30.3 & 20.3 &\first{}\textbf{61.1} & \first{}\textbf{27.2} &\first{}\textbf{25.0} &0.0 & 13.8 &\first{}\textbf{78.6} &44.4 & \first{}\textbf{66.6} & \first{}\textbf{35.5} &\first{}\textbf{30.7}  \\
 \bottomrule
\end{tabular}
}
\caption{\textbf{Comparison in Single-view accuracy~\cite{diffcad} on SUN2CAD against FoundationPose~\cite{foundationpose}.} We additionally provide GT depth maps to FoundationPose to better match their original setting. Our method still surpasses \cite{foundationpose} + GT depth in average alignment accuracies.
} 
\label{tab:tab:found_sun}
\end{table*}

\section{Additional Qualitative Results}
\label{supp:qual}

\subsection{Qualitative Results in NOC Map Prediction}
\begin{figure}[htbp]
\centerline{\includegraphics[width=249pt]{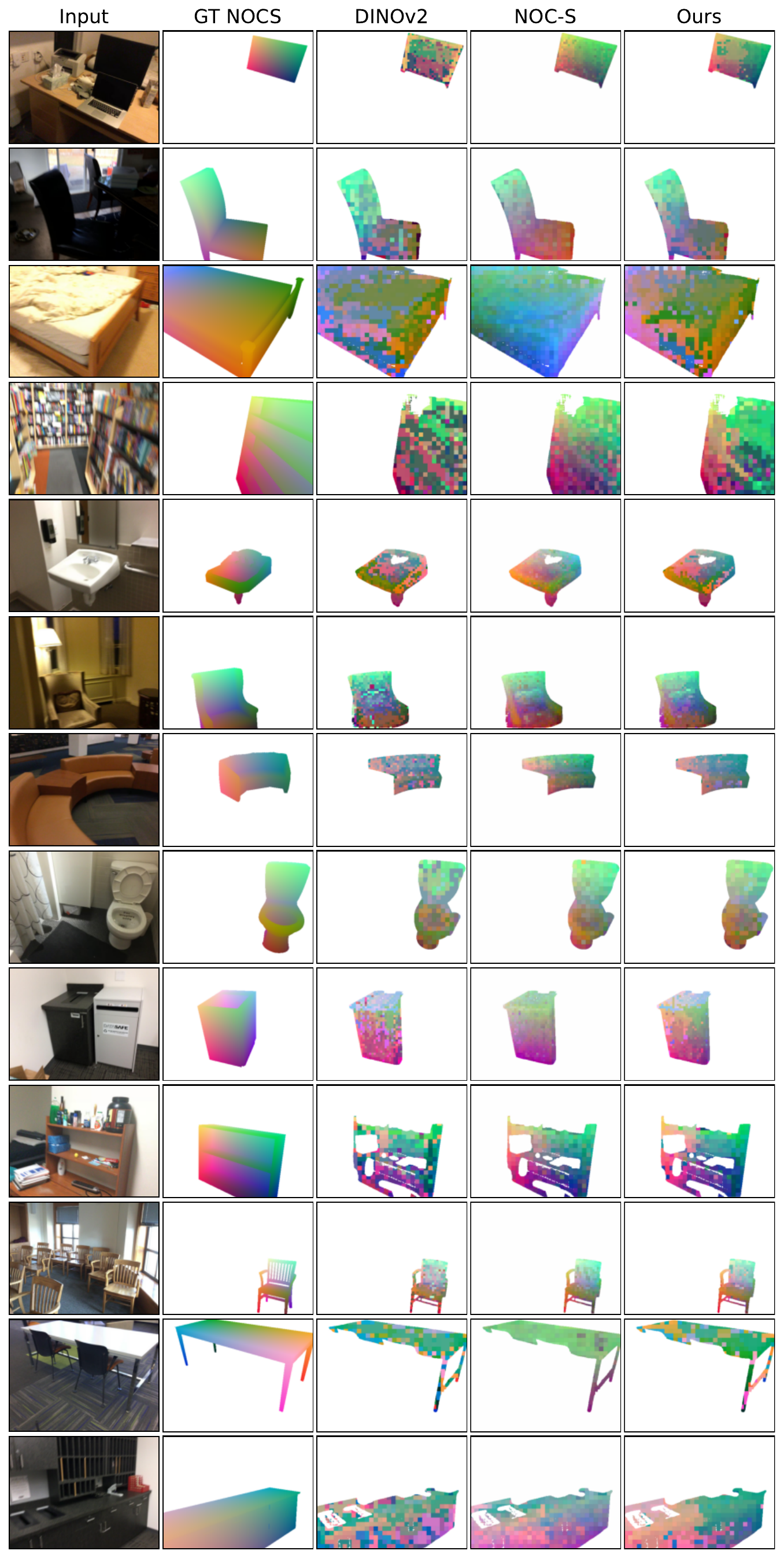}}
\caption{\textbf{Qualitative results in NOC prediction on ScanNet25k dataset.} Our solution provides better smoothness and correctness than DINOv2, while NOC-S learns to output a smooth NOC map, multiple samples show shifted or invalid NOC range compared to ground-truth NOC maps.}
\label{fig:nocresult_supp}
\end{figure}

\begin{figure}[htbp]
\centerline{\includegraphics[width=170pt]{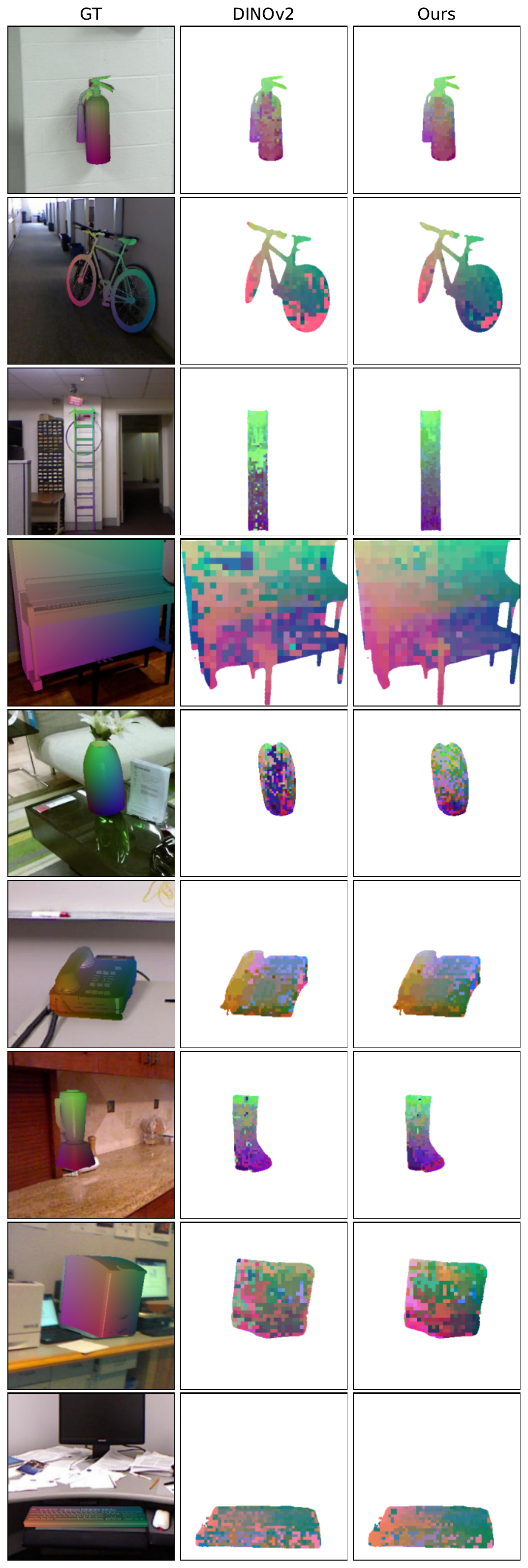}}
\caption{\textbf{Qualitative results in NOC prediction on SUN2CAD dataset.} Our solution provides better smoothness and correctness than DINOv2 in the same zero-shot setting.}
\label{fig:nocresult_supp_sun}
\end{figure}

Figure~\ref{fig:nocresult_supp} and Figure~\ref{fig:nocresult_supp_sun} show the NOC prediction results of our method and other NOC predictor baselines, NOC-S and DINOv2, on ScanNet25k~\cite{scannet} and SUN2CAD, respectively. Our method produces smoother, more accurate NOC maps, whereas DINOv2 exhibits noise due to object symmetry (e.g., left vs. right). While NOC-S produces a shifted NOC distribution in some samples, our method did not suffer from this problem due to direct correspondence matching between a given 3D model and an input image.


\subsection{Qualitative Results in Fine Pose Optimization}
\begin{figure}[]
\centerline{\includegraphics[width=236pt]{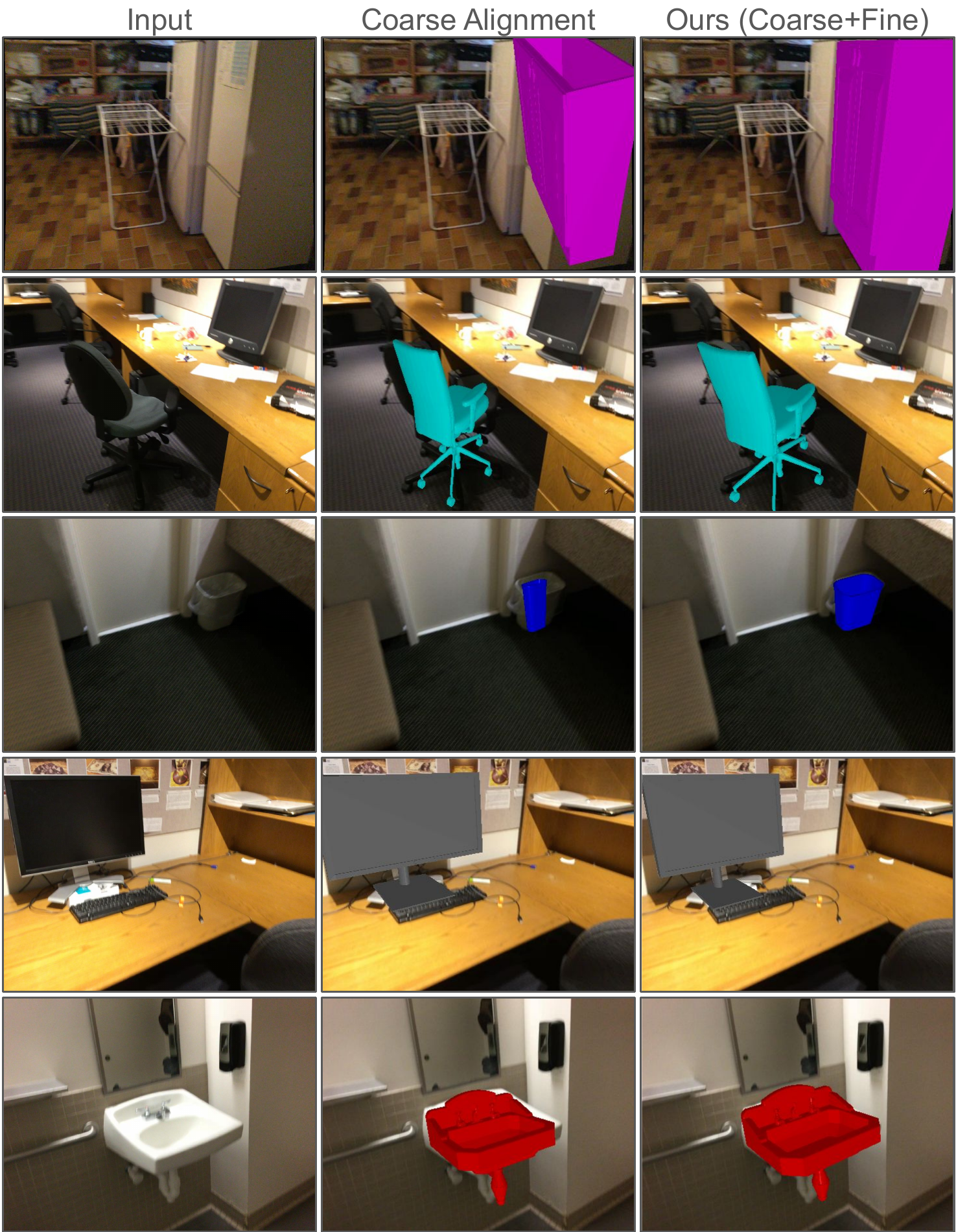}}
\caption{\textbf{Qualitative results of dense image-based alignment optimization,} shown before (coarse alignment) and after fine pose optimization.}
\label{fig:fine_opt}
\end{figure}

Figure~\ref{fig:fine_opt} illustrates a comparison between poses from our coarse alignment prediction and refined ones from our fine alignment pipeline in ScanNet25k images. The results suggest an improvement in both rotation and object placement to match the appearance of the object in the input images.

\subsection{Qualitative Results in ScanNet25k Images}
\begin{figure*}
\centering
\includegraphics[width=1\linewidth]{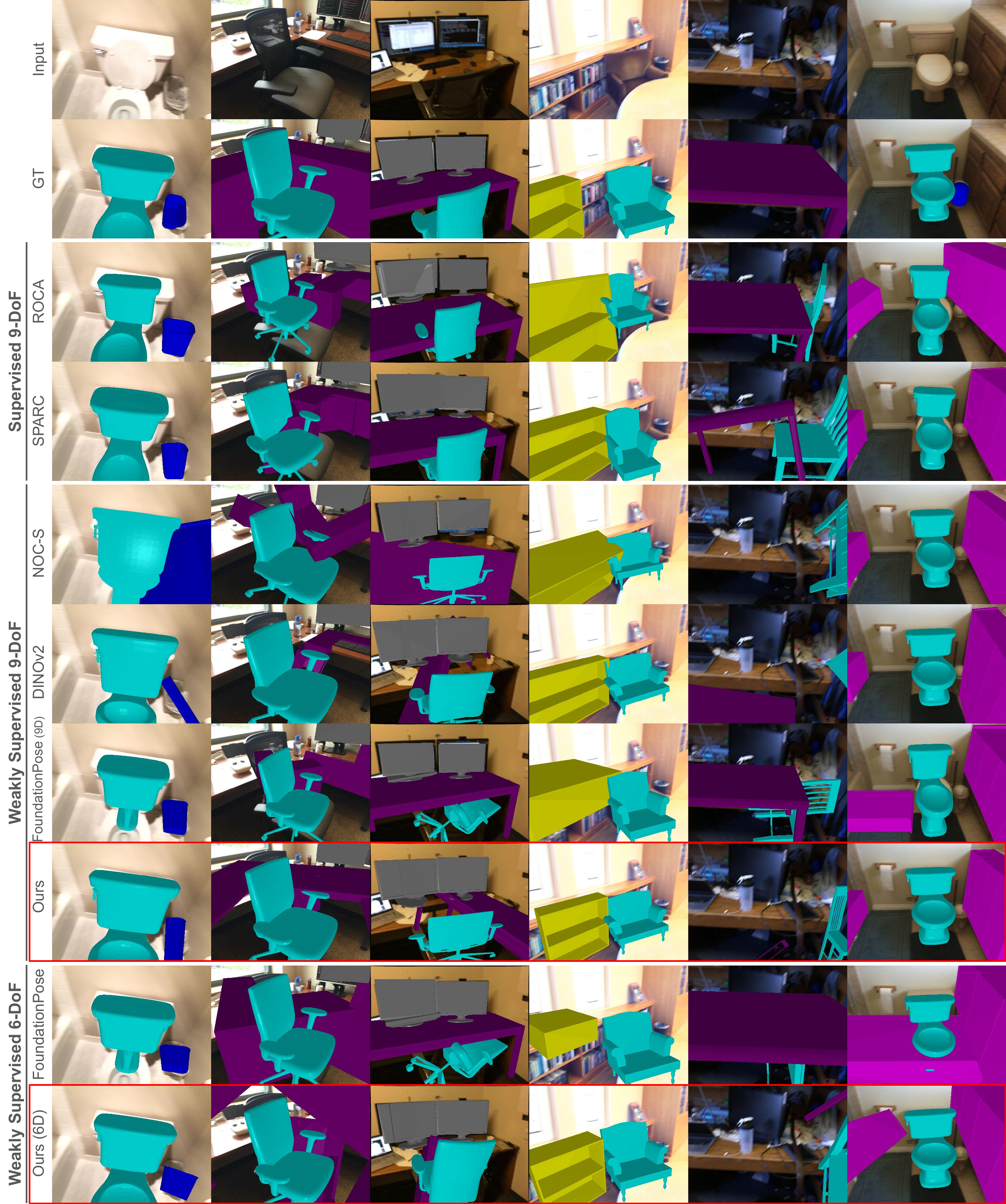}
  \caption{ \textbf{Random test samples from ScanNet25k.} We compare entire-scene CAD alignment against 9-DoF supervised methods (SPARC~\cite{sparc} and ROCA~\cite{roca}), 9-DoF weakly supervised methods (DiffCAD~\cite{diffcad}), and 6-DoF weakly supervised baselines (FoundationPose~\cite{foundationpose})}
  
\label{fig:sceneresult_p1}
\end{figure*}

\begin{figure*}
\centering
\includegraphics[width=1\linewidth]{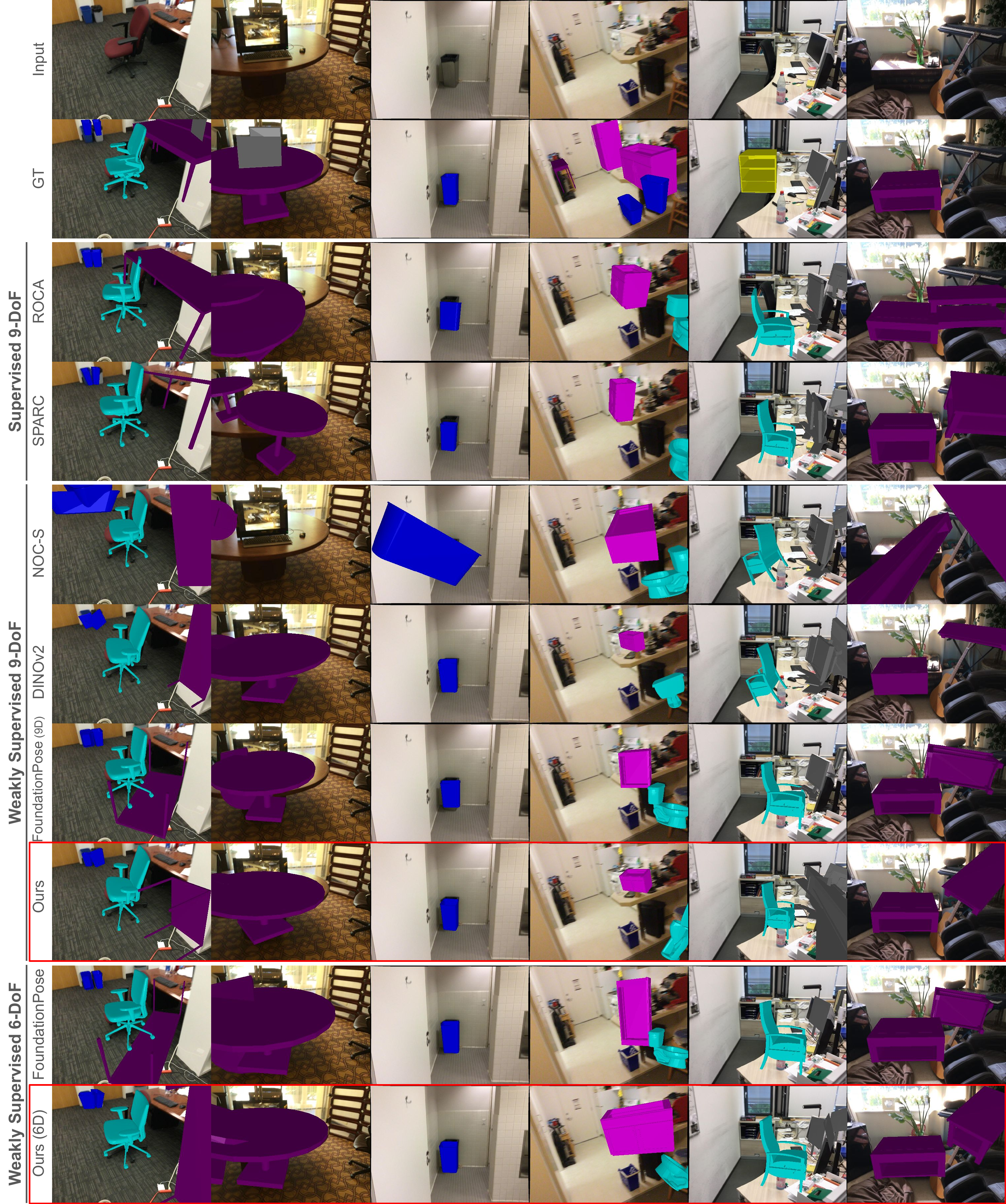}
  \caption{ \textbf{Random test samples from ScanNet25k.} We compare entire-scene CAD alignment against 9-DoF supervised methods (SPARC~\cite{sparc} and ROCA~\cite{roca}), 9-DoF weakly supervised methods (DiffCAD~\cite{diffcad}), and 6-DoF weakly supervised baselines (FoundationPose~\cite{foundationpose})}
  
\label{fig:sceneresult_p2}
\end{figure*}

\begin{figure*}
\centering
\includegraphics[width=1\linewidth]{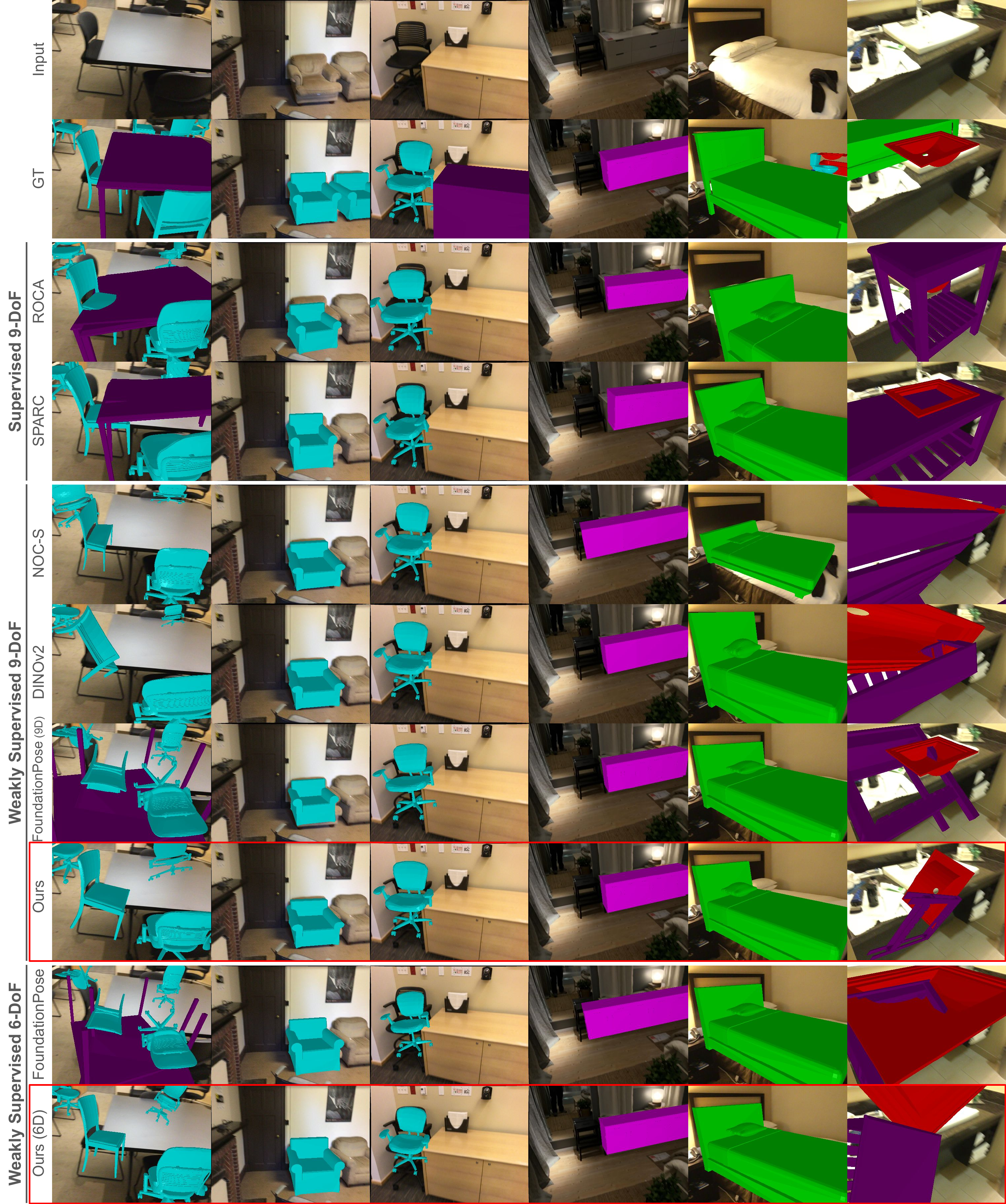}
  \caption{ \textbf{Random test samples from ScanNet25k.} We compare entire-scene CAD alignment against 9-DoF supervised methods (SPARC~\cite{sparc} and ROCA~\cite{roca}), 9-DoF weakly supervised methods (DiffCAD~\cite{diffcad}), and 6-DoF weakly supervised baselines (FoundationPose~\cite{foundationpose}).
  }
  
\label{fig:sceneresult_p3}
\end{figure*}

We present comparison results on ScanNet25K for entire-scene CAD alignment, evaluating against all our baselines: 9-DoF supervised, 9-DoF weakly supervised, and 6-DoF weakly supervised methods, with randomly selected results shown in Figure~\ref{fig:sceneresult_p1}, Figure~\ref{fig:sceneresult_p2}, and Figure~\ref{fig:sceneresult_p3}.
Please note that the 3D model in ground-truth alignments might be different from each method's retrieved 3D models.

\subsection{Qualitative Results in SUN2CAD} 
\begin{figure*}
\centering
\includegraphics[width=1\linewidth\vspace*{-0.2em}]{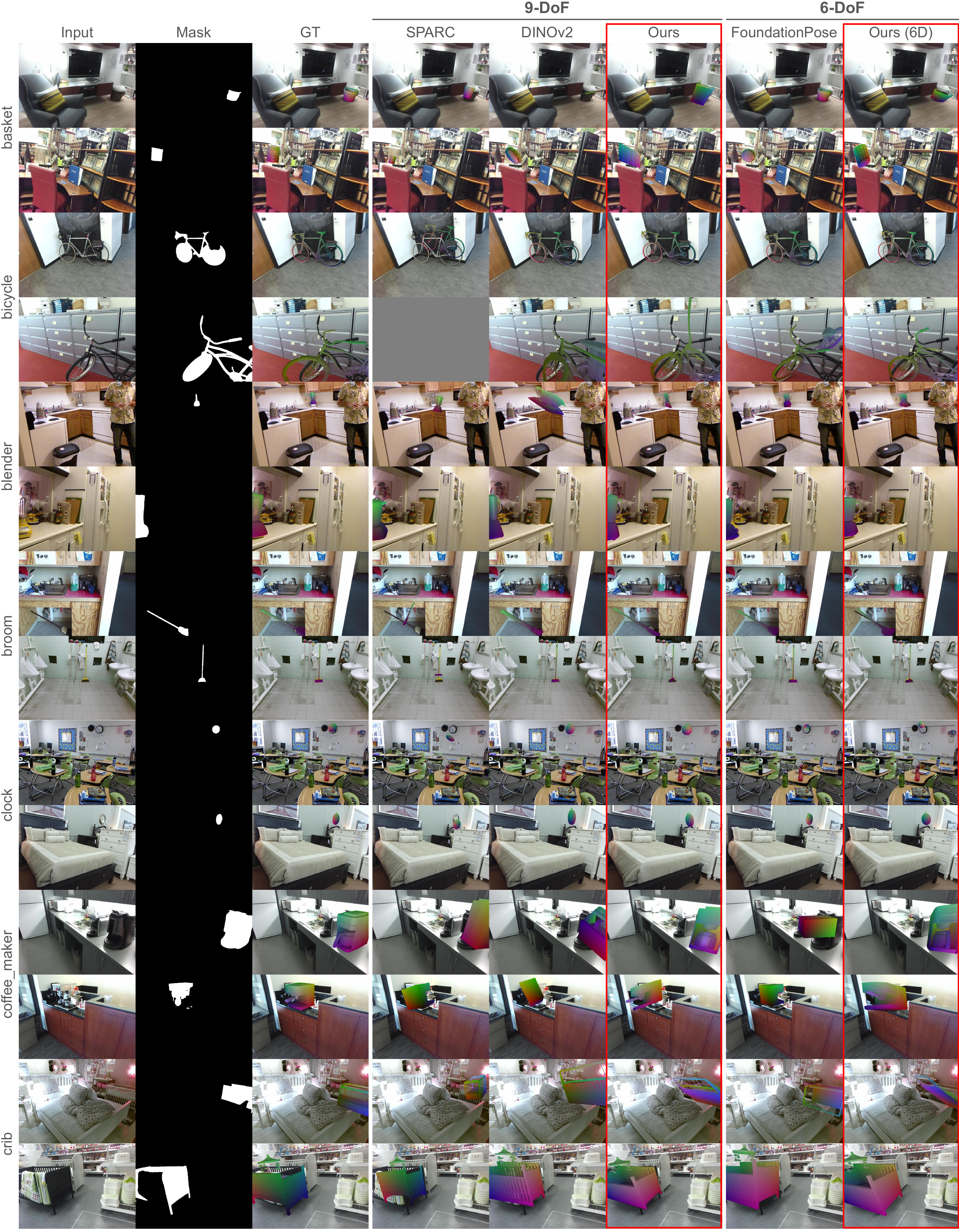}
  \caption{ \textbf{Random test samples from SUN2CAD} comparing 9-DoF pose predictions with SPARC~\cite{sparc} and DINOv2, and 6-DoF predictions with FoundationPose~\cite{foundationpose}. }
  
\label{fig:sun2cad_ran_p1}
\end{figure*}

\begin{figure*}
\centering
\includegraphics[width=1\linewidth\vspace*{-0.2em}]{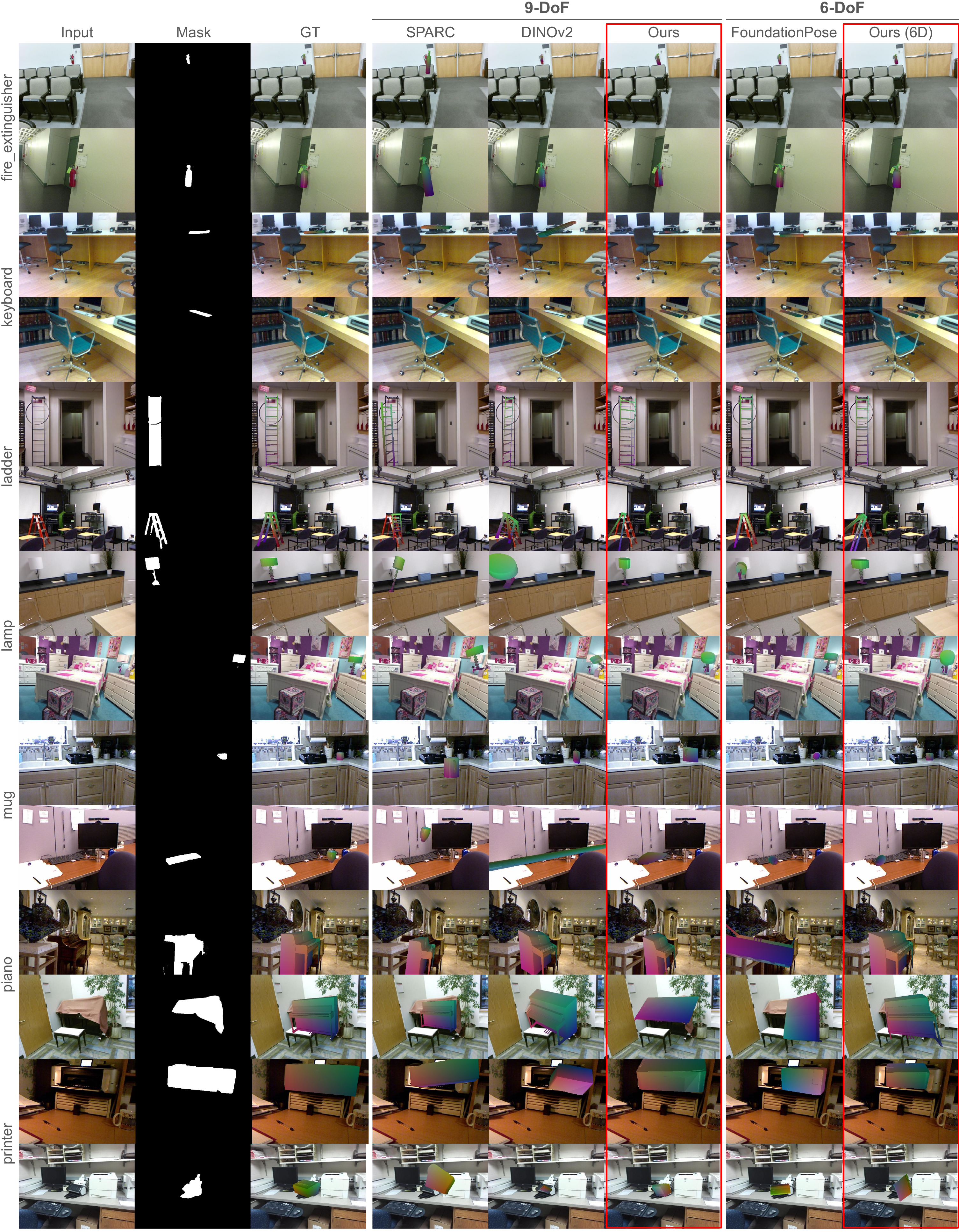}
  \caption{ \textbf{Random test samples from SUN2CAD} comparing 9-DoF pose predictions with SPARC~\cite{sparc} and DINOv2, and 6-DoF predictions with FoundationPose~\cite{foundationpose}. }
  
\label{fig:sun2cad_ran_p2}
\end{figure*}

\begin{figure*}
\centering
\includegraphics[width=1\linewidth]{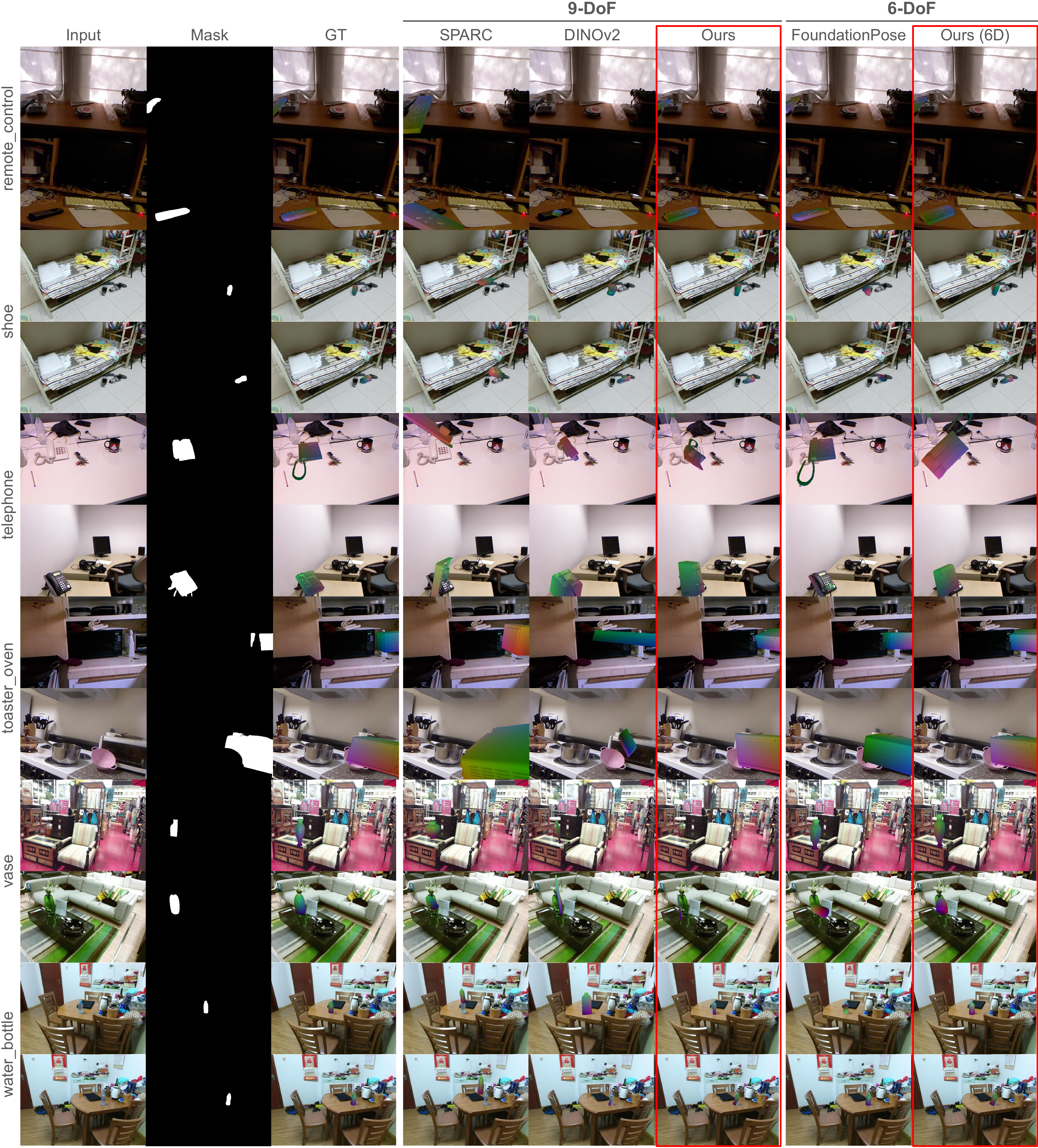}
  \caption{ \textbf{Random test samples from SUN2CAD} comparing 9-DoF pose predictions with SPARC~\cite{sparc} and DINOv2, and 6-DoF predictions with FoundationPose~\cite{foundationpose}. }
  
\label{fig:sun2cad_ran_p3}
\end{figure*}

Figure~\ref{fig:sun2cad_ran_p1}, Figure~\ref{fig:sun2cad_ran_p2}, and Figure~\ref{fig:sun2cad_ran_p3} present comparisons in CAD alignment results on random test samples from the SUN2CAD dataset between our method, SPARC~\cite{sparc}(9-DoF), DINOv2 (9-DoF) and FoundationPose~\cite{foundationpose} (6-DoF). 

\section{Additional Ablation Studies}
\label{supp:ablation}

\subsection{Study on the Architecture of Feature Adapter}
\label{supp:ab_nocarch}

We study the best network architecture for our feature adapter. We choose 3 choices: Autoencoder (AE), ViT~\cite{vit} layers, and MLP layers.
AE represents CNNs capable of decoding feature maps into pixel-level outputs, aligning with our NOC prediction objective. We follow AE architecture from Stable Diffusion~\cite{stablediffusion}'s VAE decoder.
ViT layers specialize in capturing spatial relationships through self-attention, making them well-suited for mapping feature maps to 3D structures in NOC prediction.
MLP serves as a simple yet effective linear layer for adapting zero-shot DINOv2~\cite{dinov2} features to a new feature space.

Table~\ref{tab:supp_featarch} reports NOC prediction errors (NOC error) on the ScanNet25k validation set, measured as the RMSE between predicted and ground-truth NOC maps, averaged over object pixels within the segmented mask. Following Section~\ref{sec:featureadapter}, we construct a feature voxel grid to establish 2D-3D correspondences for NOC prediction. Among the tested architectures, MLP achieves the lowest NOC error, making it the best choice.

\subsection{Study on Training Data for Feature Adapter}
\label{supp:ab_traindata}
Table~\ref{tab:training_data} presents an ablation study on the training data used for our geometry-aware adapter.  We evaluate NOC predictions by using the same feature voxel grid construction and 2D feature map with nearest-neighbor matching, while varying the adapter model used to generate the features. The results indicate that using both rendered and generated images leads to higher NOC errors than using solely rendered images due to incorrect synthetic samples, as shown in Figure~\ref{fig:augment}.

To address this issue, we propose filtering out invalid generated images using NOC prediction. Specifically, we estimate the NOC map for each generated image by applying our nearest-neighbor matching with a feature voxel grid from rendered templates, as described in Section~\ref{sec:coarse}. Here, a generated image will serve as the input feature map instead of a real image. The predicted NOC map is then compared to the ground-truth NOC map to calculate NOC error and assess whether the generated image produces features sufficiently similar to the original rendered templates. We discard generated images with an NOC error exceeding 0.20.
Training our feature adapter on filtered and rendered images results in improved performance and the lowest NOC error.
\begin{table}[]
\centering

\small
\begin{tabular}{c|c}

\toprule
\textbf{Training data}&  \textbf{NOC error} $\pm$ \textbf{SE} $\downarrow$\\ \midrule
Rendered & 0.2313 $\pm$ 0.0022
\\ 
Generated & 0.2412 $\pm$ 0.0022\\ 
Rend+Gen & 0.2330 $\pm$ 0.0023 \\ 
\textbf{Filtered Rend+Gen} & \textbf{0.2263 $\pm$ 0.0023}\\
\bottomrule
\end{tabular}
\normalsize
\caption{\textbf{Ablation study in training data of our feature adapter.}}
\label{tab:training_data}
\end{table}




\begin{table}[]
\centering
\resizebox{235pt}{!}{%
\begin{tabular}{c|c|c|c|c}
\toprule
\textbf{Features}&\textbf{\# Rendered}&\textbf{\# Augmented} & \textbf{\# Total} & \textbf{NOC error} $\pm$ \textbf{SE} $\downarrow$ \\ \midrule
\multirow{6}{*}{DINOv2}&36&-&36 & 0.2679 $\pm$ 0.0018\\
 &-&36& 36 & 0.2822 $\pm$ 0.0018\\
 &144&-& 144 & 0.2613 $\pm$ 0.0019\\
 &-&144& 144  & 0.2775 $\pm$ 0.0018\\
 &36&108& 144  & 0.2474 $\pm$ 0.0020\\
 &\textbf{36}&\textbf{252}& \textbf{288} & \textbf{0.2466 $\pm$ 0.0020}\\ \midrule 
 \multirow{6}{*}{\textbf{Ours}}&36&-& 36 & 0.2364 $\pm$ 0.0022\\
 &-&36&36 & 0.2396 $\pm$ 0.0022 \\
 &144&-&144 & 0.2335 $\pm$ 0.0022\\
 &-&144&144  & 0.2372 $\pm$ 0.0022\\
 &36&108& 144  & 0.2278 $\pm$ 0.0023\\
 &\textbf{36}&\textbf{252}& \textbf{288} & \textbf{0.2263 $\pm$ 0.0023}\\ 
 \bottomrule
\end{tabular}
    }
    \caption{\textbf{Ablation study on feature voxel grid construction.}}
    \label{tab:supp_vox}
\end{table}

\subsection{Study on the Architecture of NOC-S}
\label{supp:nocs_predictor}
We conduct an ablation study on our introduced baseline, NOC-S, to optimize its performance for a fair comparison with our method on the ScanNet25k validation set. We tested learning NOC reconstruction objective (Eq.~\ref{eq:reconloss}) on the same choice of architectures, ViT~\cite{vit} layer, MLP, and Autoencoder (AE)~\cite{stablediffusion}. 
Unlike the feature adapter’s NOC prediction, which relies on nearest-neighbor matching, NOC-S directly outputs 2D-3D correspondences from a final MLP layer.
We hypertuned each choice for optimal performance.
The results in Table~\ref{tab:supp_nocsarch} show that ViT  achieves the lowest NOC error.

Our final model consists of a single ViT layer with a one-layer MLP, mapping encoded DINOv2~\cite{dinov2} feature maps to NOC. In addition to feature maps, we also concatenate a query CAD model's global feature, which is generated by averaging CLS tokens from all CAD model's rendering templates using DINOv2. This results in a CAD-conditioned feature map with size $\mathbb{R}^{H\times W \times 2048}$.
The ViT layer has a hidden size of 2048 with 8 attention heads, while the MLP predicts 3-channel NOC outputs.
We train the model using AdamW~\cite{adamw} optimizer using a constant learning rate $5e^{-4}$ and training batch size of $150$. 

\begin{table}[]

\centering
\small
\begin{tabular}{c|c}
\toprule
 \textbf{Architecture} & \textbf{NOC error }$\pm$ \textbf{SE} $\downarrow$\\ \midrule
 AE & 0.2485 $\pm$ 0.0024 \\
 \textbf{MLP} & \textbf{0.2263 $\pm$ 0.0023}\\
 ViT & 0.2454 $\pm$ 0.0020 \\
                  
\bottomrule
\end{tabular}
\normalsize
\caption{\textbf{Ablation study in architecture choices of our feature adapter.} MLP provides the best NOCS prediction error over other choices.}
\label{tab:supp_featarch}

\end{table}

\begin{table}[]
\centering
\small

\begin{tabular}{c|c}
\toprule
 \textbf{Architecture} & \textbf{NOC error }$\pm$ \textbf{SE} $\downarrow$ \\ \midrule
 AE & 0.2094 $\pm$ 0.0015 \\
 MLP &0.2040 $\pm$ 0.0014\\
 \textbf{ViT} & \textbf{0.1966 $\pm$ 0.0017}\\
                  
\bottomrule
\end{tabular}
\normalsize

\caption{\textbf{Ablation study in architecture choices of NOC-S.} ViT provides the best NOCS prediction error.}
\label{tab:supp_nocsarch}
\end{table}

\subsection{Study on Feature Voxel Grid and Templates}
Table~\ref{tab:supp_vox} presents the impact of different template types on feature voxel grid construction on ScanNet25k validation set. Using only augmented templates (generated images) results in higher NOC errors than rendered images in the NOC output from feature matching. However, combining both rendered and generated images yields better performance. The improvements are more significant when using DINO features ($-0.0309$) than our GFS features ($-0.0109$). Nevertheless, our GFS features result in lower NOC errors than using DINOv2 features alone.

\subsection{Study on Triplet Loss Hyperparameters}
\begin{figure*}[!htbp]
\centering
\includegraphics[width=495pt]{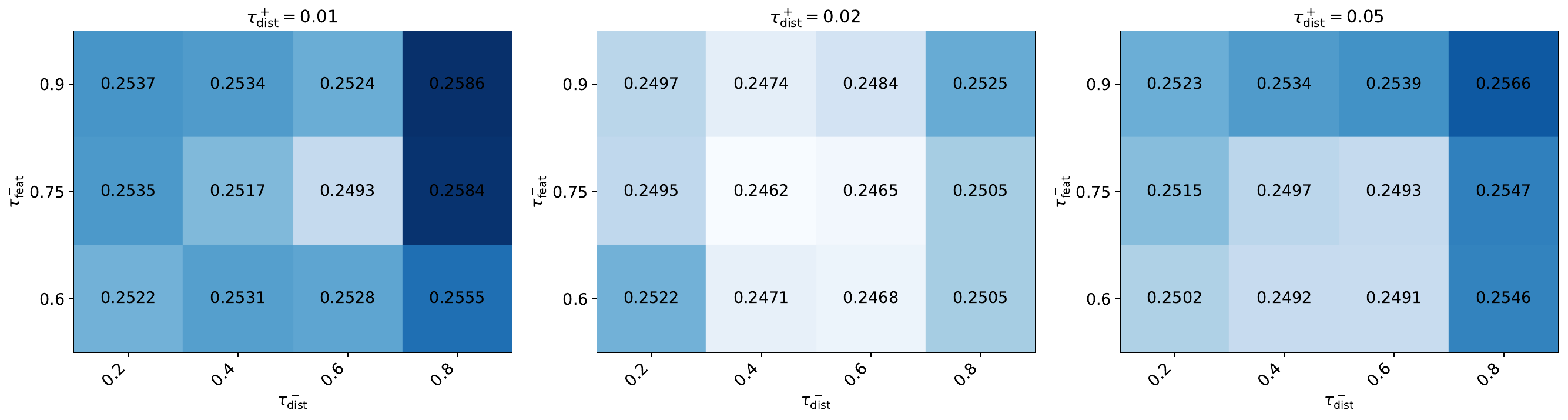}
\caption{\textbf{Ablation study in Triplet loss hyperparameters $\tau^{+}_{\text{dist}}$, $\tau^{-}_{\text{dist}}$, and $\tau^{-}_{\text{feat}}$ on NOC error metric.}}
\label{fig:triplethyperparams}
\end{figure*}

In addition to the feature adapter loss study of $\beta$ in Section~\ref{sec:expresults}, we present a hyperparameter study on $\tau^{+}_{\text{dist}}$, $\tau^{-}_{\text{dist}}$, and $\tau^{-}_{\text{feat}}$ using ScanNet25k validation set on NOC error metric.
The results are shown in Figure~\ref{fig:triplethyperparams}. Higher values of $\tau^{+}_{\text{dist}}$ result in larger NOC errors due to selecting features from different object parts as positive pairs. Similarly, excessively high values of $\tau^{-}_{\text{dist}}$ and $\tau^{-}_{\text{feat}}$ reduce negative sample diversity, increasing NOC error. Likewise, very low $\tau^{+}_{\text{dist}}$ values reduce positive pair variation, also leading to higher NOC error.

\subsection{Study on Dense Image-based Alignment Hyperparameters}

We study the single-view alignment accuracies of our dense image-based alignment hyperparameters $\lambda_{\text{NOC-A}}, \lambda_{\text{m}},$ and $ \lambda_{\text{d}}$ on ScanNet25k validation set.
To normalize each loss term, we first compute the mean alignment loss using initial poses from our coarse pose estimator. The mean values are $N_{\text{NOC-A}} = 0.33$, $N_{\text{m}} = 1$, and $N_{\text{d}} = 0.067$. Each term is then normalized relative to the smallest mean loss, yielding the scaled hyperparameters:
$\lambda_{\text{NOC-A}} = N_{\text{NOC-A}}\lambda_{\text{NOC-A}}', \lambda_{\text{m}} = N_{\text{m}}\lambda_{\text{m}}'$, and $\lambda_{\text{d}} = N_{\text{d}}\lambda_{\text{d}}'$.

\begin{figure}[htbp]
\centerline{\includegraphics[width=249pt]{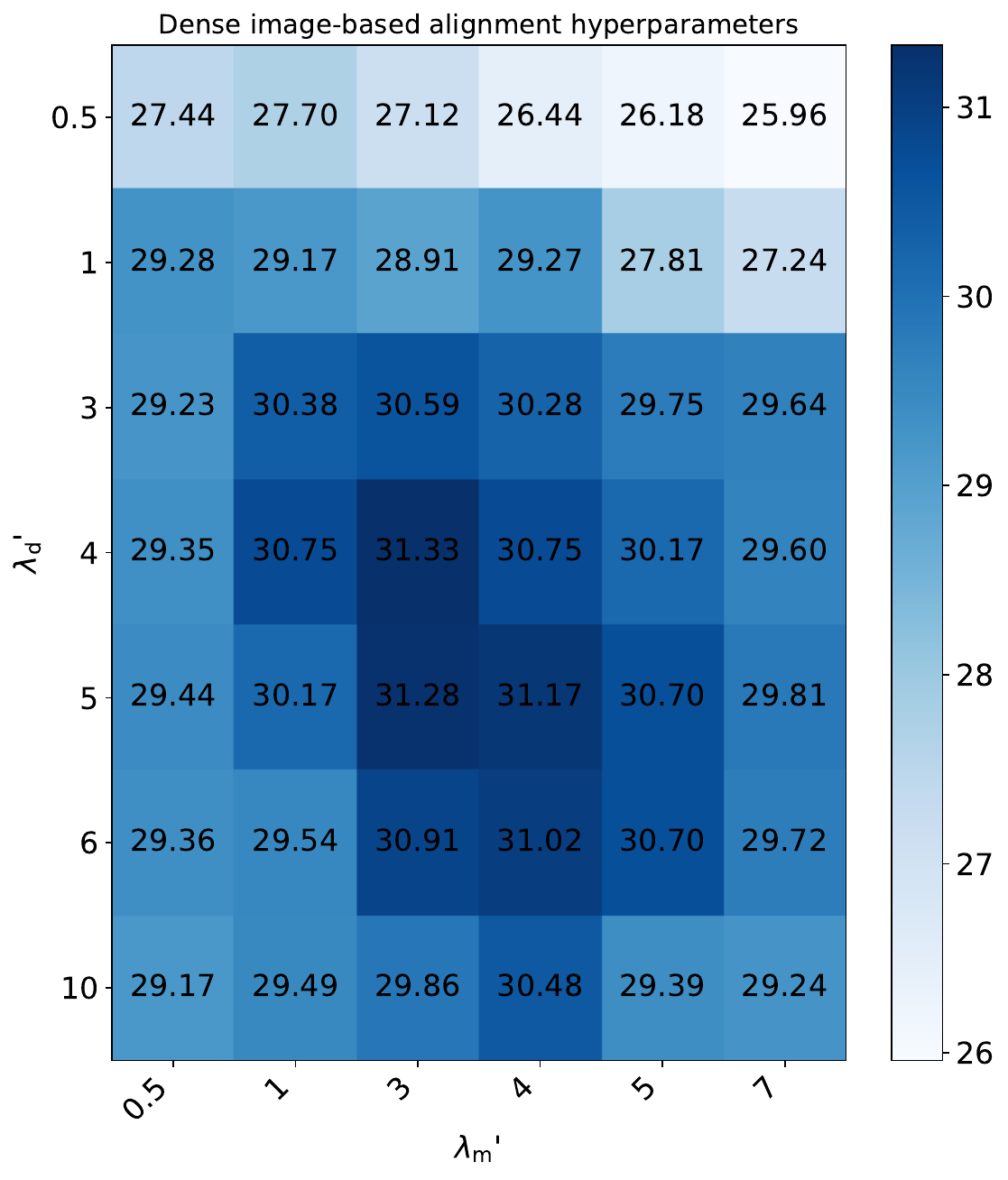}}
\caption{\textbf{Ablation study in Dense Image-based Alignment Hyperparameters on Single-view alignment accuracy~\cite{diffcad}.}}
\label{fig:dense_hyperparams}
\end{figure}

For hyperparameter tuning, we fixed $\lambda_{\text{NOC-A}}' = 1$ and perform a grid search for $\lambda_{\text{m}}'$ and $\lambda_{\text{d}}'$. As shown in Figure~\ref{fig:dense_hyperparams}, the highest alignment accuracy is achieved with $\lambda_{\text{m}}' = 3$ and $\lambda_{\text{d}}' = 4$, resulting in final values of $\lambda_{\text{NOC-A}} = 0.33$, $\lambda_{\text{m}} = 3$, and $\lambda_{\text{d}} = 0.27$.

\section{Failure Cases}
\label{supp:failurecase}

\begin{figure}[]
\centerline{\includegraphics[width=236pt]{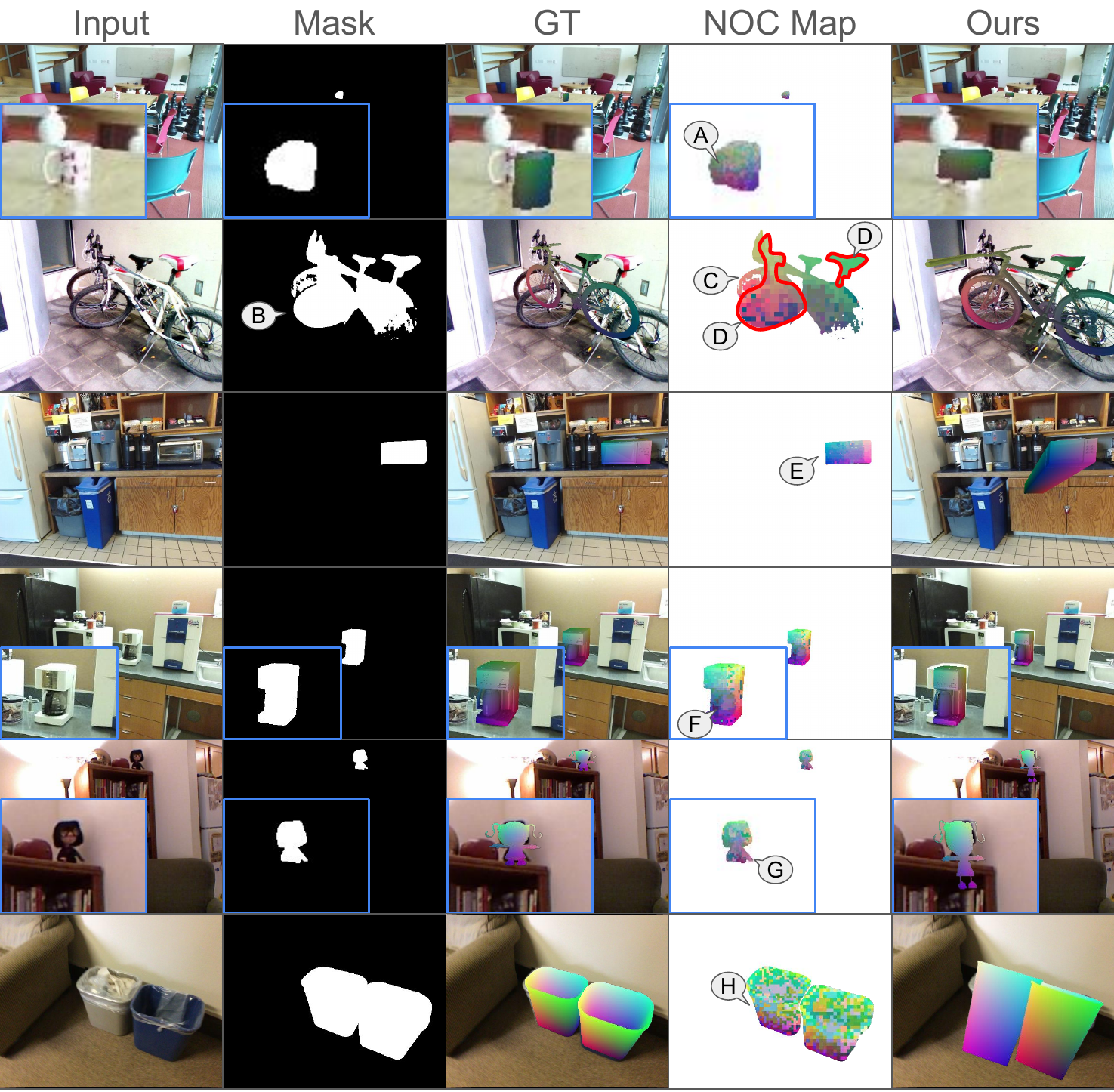}}
\caption{\textbf{Failure cases;} (A) Poor input image quality; (B) incorrect input mask, causing two objects' features to blend together (C, D) for CAD alignment; (E) lack of edge cues for inferring object depth; (F, G) alignment ambiguity; (H) Thin object.}
\label{fig:fail}
\end{figure}

Figure~\ref{fig:fail} illustrates failure cases. Poor input image quality, such as small objects (A), degrades DINOv2 features, making it difficult to distinguish object parts correctly, which can lead to incorrect pose estimation.

Failures in dependencies, such as depth prediction or mask segmentation, also affect our method. For example, in (B), the mask prediction merges two bicycles into one. As a result, while our method produces a good NOC map for the target bicycle in the back (C), it also unintentionally includes parts of the front bicycle (D), causing misalignment by aligning with the front bicycle instead.

Objects with minimal edge cues, such as a toaster oven with severe self-occlusion, present another challenge. The NOC map (E) lacks sufficient edge information to infer depth, leading to an incorrect shape prediction.

Alignment ambiguity arises when multiple matched parts have inconsistent spatial relationships between the CAD model and the real object. Our method aims to cover all local correspondences, whereas humans may prioritize specific object parts. In (F), the CAD model of a coffee maker has a kettle on the left, whereas the real object has it in the middle. Our method prioritizes aligning the kettle position rather than focusing on the overall coffee maker shape, deviating from the ground-truth. Similarly, in (G), when aligning a CAD model with significant visual differences from its target object, our method tries to match individual parts, such as the doll's hand, while a human might prioritize aligning the doll's head. 

Lastly, although (H) clearly distinguishes between top and bottom, the thin structure complicates distinguishing inside from outside on the same side, making it more prone to failure compared to other objects.

\section{Societal Impacts}
\label{supp:societal}
Our work on 9-DoF pose estimation benefits real-world applications in synthetic environments such as VR and gaming, where safety is not a concern. However, its accuracy may be insufficient for safety-critical tasks like autonomous driving. Reliance on predicted depth alone can introduce translation-scale ambiguity, and further accuracy improvements are needed to mitigate potential risks in high-stakes environments.

\end{document}